\documentclass{article}

\usepackage{microtype}
\usepackage{graphicx}
\usepackage{subfigure}
\usepackage{booktabs} % for professional tables
\usepackage{soul}
\usepackage{amsmath}

\usepackage{amsfonts} 
\usepackage{enumitem}
\usepackage{graphicx}
\usepackage{graphbox}
\usepackage{float}
\usepackage{dblfloatfix}
\usepackage{nicematrix}
\usepackage{siunitx}

\usepackage{hyperref}

\usepackage[accepted]{icml2023}

\usepackage{amsmath}
\usepackage{amssymb}
\usepackage{mathtools}
\usepackage{amsthm}

\usepackage[capitalize,noabbrev]{cleveref}

\theoremstyle{plain}

\theoremstyle{definition}

\theoremstyle{remark}

\icmltitlerunning{Curious Replay for Model-based Adaptation}

\begin{document}

\twocolumn[
\icmltitle{Curious Replay for Model-based Adaptation}

% It is OKAY to include author information, even for blind
% submissions: the style file will automatically remove it for you
% unless you've provided the [accepted] option to the icml2023
% package.

% List of affiliations: The first argument should be a (short)
% identifier you will use later to specify author affiliations
% Academic affiliations should list Department, University, City, Region, Country
% Industry affiliations should list Company, City, Region, Country

% You can specify symbols, otherwise they are numbered in order.
% Ideally, you should not use this facility. Affiliations will be numbered
% in order of appearance and this is the preferred way.
\icmlsetsymbol{equal}{*}

\begin{icmlauthorlist}
\icmlauthor{Isaac Kauvar}{equal,stanford}
\icmlauthor{Chris Doyle}{equal,stanford}
\icmlauthor{Linqi Zhou}{stanfordcs}
\icmlauthor{Nick Haber}{stanford}
\end{icmlauthorlist}

\icmlaffiliation{stanford}{Graduate School of Education, Stanford University, Stanford, CA, USA}
\icmlaffiliation{stanfordcs}{Department of Computer Science, Stanford University, Stanford, CA, USA}

\icmlcorrespondingauthor{Isaac Kauvar}{ikauvar@stanford.edu}
\icmlcorrespondingauthor{Nick Haber}{nhaber@stanford.edu}

% You may provide any keywords that you
% find helpful for describing your paper; these are used to populate
% the "keywords" metadata in the PDF but will not be shown in the document
\icmlkeywords{Machine Learning, ICML}

\vskip 0.3in
]

% this must go after the closing bracket ] following \twocolumn[ ...

% This command actually creates the footnote in the first column
% listing the affiliations and the copyright notice.
% The command takes one argument, which is text to display at the start of the footnote.
% The \icmlEqualContribution command is standard text for equal contribution.
% Remove it (just {}) if you do not need this facility.

%\printAffiliationsAndNotice{}  % leave blank if no need to mention equal contribution
\printAffiliationsAndNotice{\icmlEqualContribution} % otherwise use the standard text.

\begin{abstract}
Agents must be able to adapt quickly as an environment changes. We find that existing model-based reinforcement learning agents are unable to do this well, in part because of how they use past experiences to train their world model.
Here, we present Curious Replay---a form of prioritized experience replay tailored to model-based agents through use of a curiosity-based priority signal. Agents using Curious Replay exhibit improved performance in an exploration paradigm inspired by animal behavior and on the Crafter benchmark. DreamerV3 with Curious Replay 
surpasses state-of-the-art performance on Crafter, achieving a mean score of 19.4 that substantially improves on the previous high score of 14.5 by DreamerV3 with uniform replay, while also maintaining similar performance on the Deepmind Control Suite. Code for Curious Replay is available at \href{https://github.com/AutonomousAgentsLab/curiousreplay}{github.com/AutonomousAgentsLab/curiousreplay}.

\end{abstract}

\section{Introduction}
Change is unavoidable. Robust artificially intelligent (AI) agents must be capable of quickly adapting to changing circumstances. In the face of novel states and shifting conditions, agents---whether self-driving cars, home-assistant robots, or financial decision makers---must effectively update their understanding of the world and their policies for acting in it. Animals have evolved to skillfully contend with the challenges of changing environments. Can they serve as an existence proof and source of inspiration for synthesizing such flexible intelligence?   

Consider, for instance, a simple change in one's environment: the appearance of a new object. Animals, from rodents to primates, will often adapt to such a change by first investigating the object \cite{glickman1966curiosity}. This is an effective strategy as it addresses a key aspect of adaptation: assembling data to update models that guide behavior. In this work, we investigate such functionalities of gathering and utilizing information about environmental change in the context of model-based deep reinforcement learning.

\begin{figure}[t]
\vskip 0.2in
\begin{center}
\centerline{\includegraphics[width=\columnwidth]{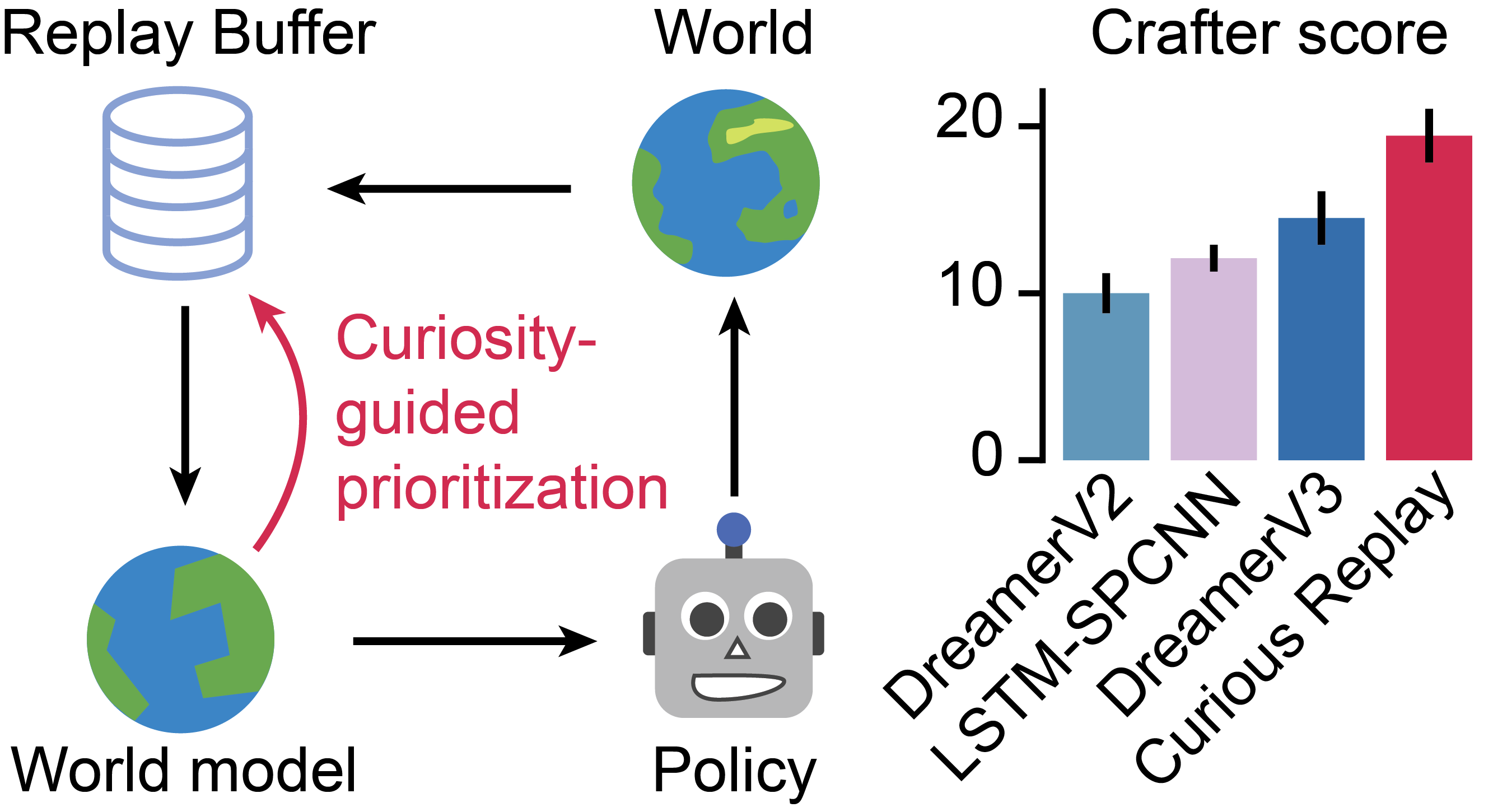}}
\vskip -0.1in
\caption{Curious Replay closes the loop between experience replay and world model performance  by using curiosity-guided prioritization to promote training on experiences the model is least familiar with. Curious Replay improves the adaptability of model-based agents and yields a new state-of-the-art score on Crafter.}
\label{fig1}
\end{center}
\vskip -0.4in
\end{figure}

We identify surprising deficiencies in the adaptability of a state-of-the-art model-based agent (Dreamer, with Plan2Explore for intrinsically-motivated settings \cite{hafner2020mastering, sekar2020planning, hafner2023mastering}). In an object interaction assay, we find that---unlike animals--- Plan2Explore agents do \textit{not} quickly interact with a novel object. Moreover, in a nonstationary variant of the Deepmind Control Suite, DreamerV2 does not adapt well. 
These results reveal shortcomings in the Dreamer agent that may broadly impact its ability to achieve its full potential in challenging settings.

We find a root cause of Dreamer's poor adaptability: reliance on uniform sampling of its experience replay buffer.
We address this problem with Curious Replay---a new approach to prioritizing replay buffer sampling when updating an agent's models---and demonstrate its profound benefits for adaptation of model-based agents, including achievement of a new state-of-the-art on Crafter (Figure \ref{fig1}).

\begin{figure*}[b]
\centerline{\includegraphics[width=\textwidth]{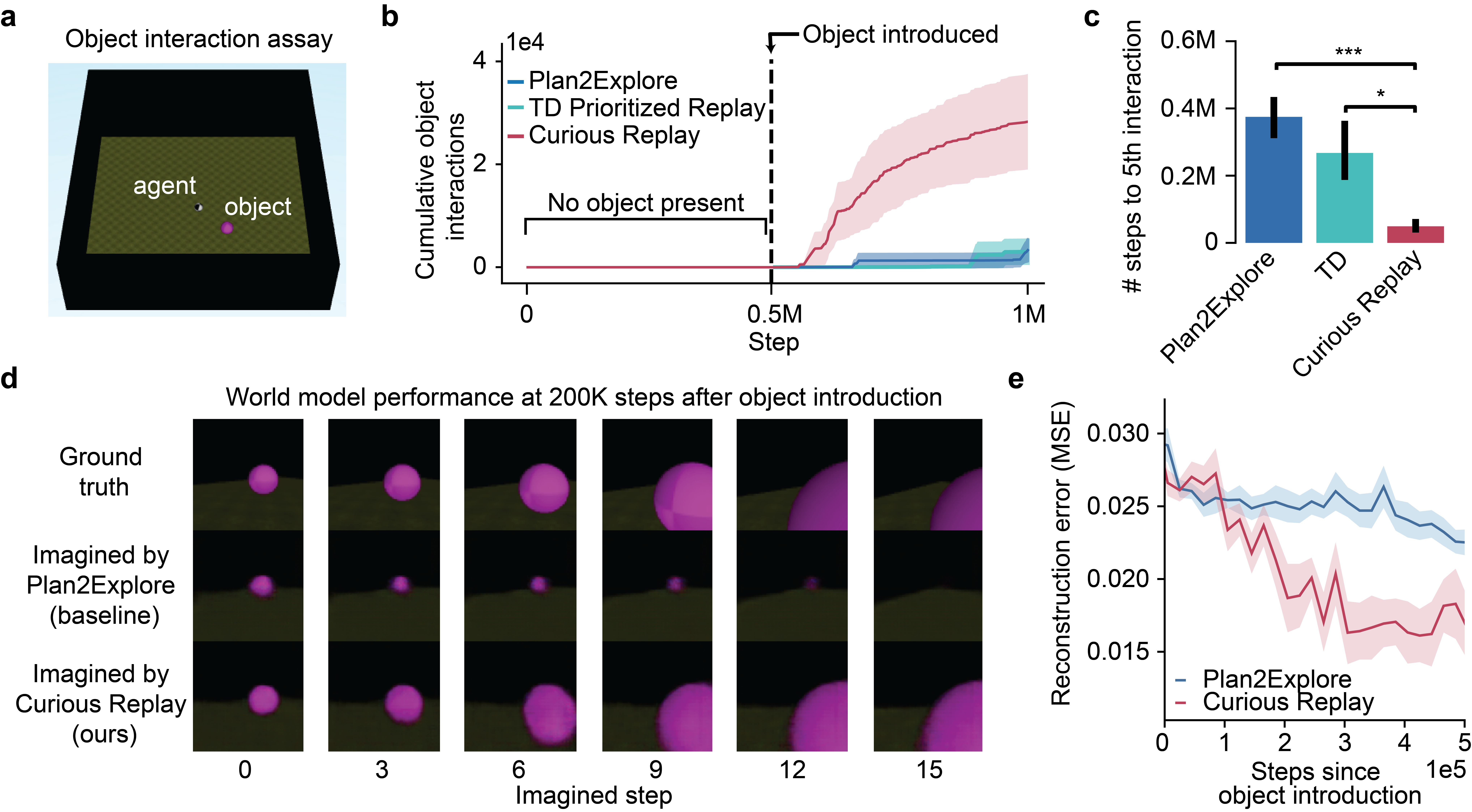}}
% \vskip -0.15in
\caption{Inspired by animal behavior, we investigate adaptation using an object interaction assay. a) In this 3D physically-simulated assay, an intrinsically-motivated agent can explore an empty arena for 500K steps, at which point a novel object is introduced. The expectation is that agents, like animals, will quickly begin to interact with the novel object. b) The baseline Plan2Explore agent does not quickly interact with the object. Curious Replay dramatically improves the adaptability of Plan2Explore. c) Curious Replay yields significantly faster time-to-interaction than both Plan2Explore and Plan2Explore with prioritized experience replay using value-based temporal-difference error (TD) prioritization  \cite{schaul2015prioritized} (n=7 each condition, independent t-test with fdr-bh correction). d) Plan2Explore fails to quickly model the novel object (visualized by an example egocentric imagined rollout), even though it has substantial experience observing the object (see Figure \ref{fig_npix_magenta}). In contrast, the Curious Replay agent quickly learns to accurately model the object. Each displayed rollout is from the lowest-error model (at 200K steps after object introduction) of 7 random seeds. e) Summary of model performance over time, demonstrating the faster rate at which Curious Replay learns to accurately model the object (n=7 each, mean +/- sem).}
\label{fig2}
\vskip -0.15in
\end{figure*}

\pagebreak

Curious Replay utilizes curiosity as an intrinsic  signal --- not for action selection, but to choose experiences for model updates. This greatly boosts performance in changing environments while maintaining similar performance in unchanging ones. Moreover,  Curious Replay also offers benefits in environments where the need for adaptation is less obvious, like the open-world game Crafter \cite{hafner2021crafter}, indicating its potential value for broadly improving model-based agents.

The success of Curious Replay stems from the idea that for a model-based system to adapt effectively, the model must keep pace with changes in the environment. This is crucial as an inaccurate model can hinder appropriate action selection, especially for actions related to new environmental changes, creating a compounding issue where poor actions lead to poor data collection. To overcome this, our approach is to emphasize training the model on unfamiliar or challenging aspects of the environment. This differs from uniform replay buffer sampling, which can lead to the model spending too much time training on old or irrelevant experiences, and neglecting important updates to its world model.

Curious Replay expands on the established success of prioritized experience replay \cite{schaul2015prioritized} by tailoring it for the model-based setting with curiosity, yielding large improvements in agent adaptability.

In sum, we make the following key contributions:
\begin{itemize}[noitemsep,nolistsep]
    \item We describe Curious Replay (CR), a method that aids model-based RL agent adaptation by prioritizing replay of experiences the agent knows the least about.
    \item We introduce assays for studying RL in changing environments, including an object interaction task inspired by animal behavior, and we find that DreamerV2 agents fail to quickly adapt to the changing environments. 
    \item We show that CR improves performance in these adaptation assays, with more than $6\times$ improvement at object interaction. 
    \item We show that combining CR with DreamerV3 yields a new state-of-the-art score on Crafter, improving upon DreamerV3 by a factor of 1.33, and while maintaining a similar overall score on Deepmind Control Suite. 
\end{itemize}

\section{Changing environments}
\subsection{RL in changing environments}
\label{sec:changing}
In reinforcement learning (RL), agents experience an environment as a sequence of states and choose actions for each state to maximize expected reward. The problem can be formulated as a partially observed Markov decision process (POMDP) parameterized by the tuple $(S, A, T, R, \Omega, O, \gamma)$. Here, $S$ is the set of states, which are not directly accessible to the agent, $A$ is the set of actions, $T$ is the action-dependent transition probabilities between states, $R$ is the reward function, $\Omega$ is the set of observations, $O$ is the function that transforms the state to an observation $x \in \Omega$, and $\gamma \in (0, 1)$ is the discount on future rewards. We consider image observations $x \in \mathbb{R}^{M \times N \times C}$, with dimensions $M$ and $N$ and color channels $C$, and both discrete- and continuous-action space environments. Rewards $r \in \mathbb{R}$ may depend on components of the state related to the agent (intrinsic rewards) or the environment (extrinsic rewards). The objective is to learn a policy $\pi$ that maximizes $\mathbb{E}_\pi[\sum_{t \ge  0} \gamma^t r_t]$.

We focus on changing environments, in which there can be a step change in the dynamics or states. Changing environments are a special case of the POMDP where there is a natural representation that separates a discrete latent variable $\chi$ from $S$ to represent phases of the environment. In this framing, $T$, $R$, $\Omega$ and $O$ all become functions of the unobserved variable $\chi$, and $T$ defines both the next state and the next phase of the environment. This framing is natural for certain environments characterized by important step changes. These can occur as a result of factors external to the agent (e.g. at time $t=T_0$, a light in the environment is programmed to turn on), or because of achievements by the agent (e.g. the agent enters a new area). A consequence of such a change is that aspects of the agent's model are no longer accurate or comprehensive. Observations may now be novel, or actions may now have a different effect, and the agent must adapt.

Such changing environments differ from benchmarks such as Deepmind Control Suite \cite{tassa2018deepmind}, where the environment stays consistent throughout learning. They also differ from the task of training across multiple environments followed by evaluation on a test environment, with the agent explicitly cued that it is in the test phase \cite{parisi2021interesting}. Moreover, we do not focus on leveraging a phase of unsupervised exploration to inform downstream task-completion in a single unchanged environment \cite{sekar2020planning, laskin2021urlb}. Nor do we focus on the task of robustification from distracting stimuli, for which data augmentation has been successfully applied \cite{deng2022dreamerpro, yarats2020image}. Rather, we are interested in scenarios where some aspect of the environment changes---whether the dynamics, the observations, or the possible results of actions---and where no explicit cue tells the agent when a change happens.

We investigate adaptation of model-based agents in three settings: an intrinsically-motivated object interaction assay, variants of the Deepmind Control Suite, and Crafter \cite{hafner2021crafter}. While reminiscent of existing settings with a reward-free exploration phase followed by a task adaptation phase \cite{parisi2021interesting, laskin2021urlb, sekar2020planning}, our assays are distinct in testing how agents respond to changes in environment states or dynamics, with the agent not explicitly cued when entering a new phase.

\subsection{Object interaction assay}
The object interaction assay is inspired by animal behavior. In response to the appearance of a novel object, animals will quickly begin to investigate the object \cite{glickman1966curiosity, ahmadlou2021cell}. Notably, animals interact with the object more quickly if it represents a change to the environment (i.e. if it appears after the animal has had time to explore the object-less environment). 
We verified these phenomena with animal experiments in Figure \ref{figS1}. 

We sought to assess the investigatory behavior of AI agents in an analogous setting. We implemented the object interaction assay in a 3D egocentric, image-based environment that extends the dm\_control simulation framework \cite{tunyasuvunakool2020}. In the assay, an agent explores an empty, square arena with black walls. At $T_0=500\text{K}$ steps, a magenta ball appears near the center of the arena. The ball is stationary but untethered, and the agent can collide with it to move it around the arena. The agent has not previously experienced magenta. We also implemented an unchanging version of the assay, with the ball present from the start. 

There is no extrinsic reward in this assay, and agents must be guided by an intrinsic reward signal. Because interacting with the ball yields very different observations and dynamics than any other part of the environment, we expect that good exploratory behavior should involve interaction with the ball. We thus quantify exploration by the number of agent-ball interactions (e.g. collisions). We also quantify the accuracy of the learned world model on test episodes, as in Figure \ref{fig2}.

\begin{algorithm*}[]
    \caption{Curious Replay}
    \label{alg:example}
    \begin{algorithmic}%[1]
        \STATE {\bfseries Input:} Replay buffer $R$ that uses a SumTree structure to store the priority $p_i$ of each transition
        \STATE {\bfseries Hyperparameters:} $c$, $\beta$, $\alpha$, $\epsilon$, environment steps per train step $L$, batch size $B$, maximum priority $p_{\text{MAX}}$
        \FOR{iteration 1, 2, \ldots }
            \STATE Collect $L$ transitions $(x_t, a_t, r_t, x_{t+1})$ with policy
            \STATE Add transitions to replay buffer $R$, each with priority $p_i \leftarrow p_{\text{MAX}}$ and visit count $v_i \leftarrow 0$
            \STATE Sample batch of $B$ transitions from $R$ using probability for selecting transition $i$ as $p_i / \sum_{j = 1}^{|R|}{p_j}$
            \STATE Train world model and policy using batch, and cache loss $\mathcal{L}_{i}$ for each transition in batch
            \FOR{transition $i$ in batch}
               % \STATE Calculate loss $\mathcal{L}_{i}$
               \STATE $p_i \leftarrow c\beta^{v_{i}} + (|\mathcal{L}_{i}| + \epsilon)^\alpha$ \mbox{\quad (See equation \ref{p_eqn})}
               \STATE $v_{i} \leftarrow v_{i}+1$
            \ENDFOR
        \ENDFOR
   \end{algorithmic}   
\end{algorithm*}

\subsection{Deepmind Control Suite}

To investigate agent adaptation in an extrinsically-rewarded, changing environment, we modify the Deepmind Control Suite \cite{tassa2018deepmind} to make it a changing environment. In the Constrained Control Suite variant, we modify the cheetah, cup, and cartpole environments to add constraints on the motion of the agent's appendages (Figure \ref{fig_cdmc}). These constraints are released at $T_0=500\text{K}$ steps, necessitating adaptation to the newly available action outcomes. The agent can only achieve maximum extrinsic reward in the unconstrained phase. To adapt effectively, it must respond to the changed constraints and actively investigate the newly available states. In the Background-Swap Control Suite, we leverage the Distracting Control Suite framework \cite{stone2021distracting}, using one static image as a background until $T_0=1$M, at which point the background changes to a new image. At $T_1=2$M, the background reverts to the original image. We can thus test adaptation at $T_0$, and maintenance of performance from the first environment phase at $T_1$. We additionally assess model performance on the original, unchanging, Deepmind Control Suite.

\subsection{Crafter}
We also sought to test adaptation in a more complex and well-validated setting. For this, we turned to Crafter \cite{hafner2021crafter}, a procedurally generated, open-world survival game in which an agent pursues a variety of hierarchical achievements. Uncovering an achievement changes the states and achievements available to an agent, and agents must leverage this new knowledge and adapt subsequent behavior. For example, whereas the Collect Drink achievement has no prerequisities, Collect Coal is only unlocked by preceding achievements of Collect Wood, Place Table, and Make Wood Pickaxe. In this manner, the available states change as the agent progresses. The score reflects achievement success across trials. Humans score around 50\%.

\section{Curious Replay}
\begin{figure*}[t]
\vskip 0.2in
\begin{center}

\centerline{\includegraphics[width=\textwidth]{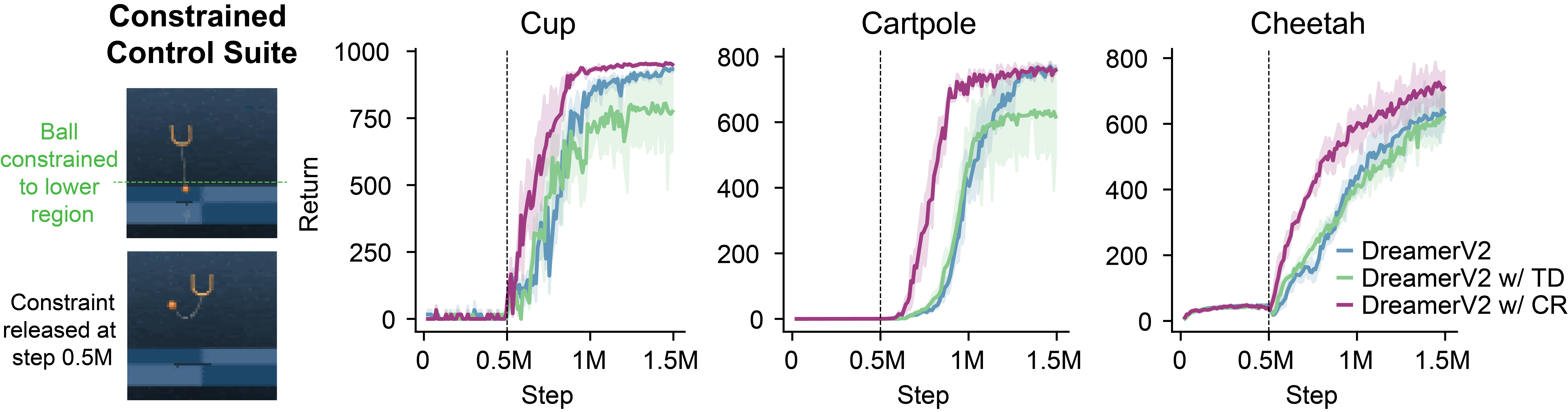}}
\vskip -0.1in
\caption{DreamerV2 w/ Curious Replay outperforms DreamerV2 and DreamerV2 w/ TD in the Constrained Control Suite (n=6 per method, mean +/- s.e.m.)}
\label{cdmc}
\end{center}
\vskip -0.4in
\end{figure*}

\subsection{Model-based RL in changing environments}
 Model-based RL leverages a state-predictive world model to inform learned action selection \cite{moerland2023model}. At least two key advantages are promised relative to model-free RL: more data-efficient training due to the compression of experience into a predictive world model, and more effective
policies that can leverage the world model for planning. Recent success with model-based agents has yielded key gains relative to model-free systems, including in data-efficiency and overall task performance \cite{schrittwieser2020mastering, hafner2020mastering, hafner2023mastering}, and we restrict our investigation here to improving model-based systems. 

Dreamer is a particularly successful architecture for model-based RL from images \cite{hafner2019dream, hafner2020mastering, hafner2023mastering, wu2022daydreamer}, with three key components: (1) a replay buffer of stored experience (2) a world model that embeds observations to a latent state and uses a forward-dynamics Recurrent State Space Model \cite{hafner2019learning} to imagine action-conditioned future states, and (3) an actor-critic policy trained on trajectories imagined by the world model. Dreamer can be effectively augmented with intrinsic reward, such as disagreement in Plan2Explore \cite{sekar2020planning}. 

Training Dreamer consists of three main steps that are cycled until convergence: (1) environment interaction, (2) world model learning, and (3) policy learning. Interaction with the environment is recorded into an experience replay buffer as sequences of observations. The world model is optimized to fit the data in the replay buffer, using uniformly sampled sequences. The actor-critic policy is then trained on imagined trajectories that are simulated by the world model, seeded at initial states recalled from the replay buffer.

Dreamer's world model allows the policy to learn using multi-step predictions in a compact latent space, yielding more sample-efficient training of sophisticated policies. 
The key drawback, however, is that if the world model is too inaccurate, the policy will be ineffective. 
This can be particularly problematic in changing environments if the model does not keep up-to-date with observed changes.

Initial object interaction experiments demonstrate that Dreamer suffers from this problem in changing environments. As shown in Figure \ref{fig2}, the baseline Plan2Explore agent struggles to represent a novel object that appears in the playground environment, even though it has observed the object for tens of thousands of steps (Figure \ref{fig_npix_magenta}). Moreover, there is a counterintuitive deficiency in the agent's behavior: the agent is ten times slower to interact with the object in the changing version of the assay than in the unchanging version (Figure \ref{figS0}). This ordering is the opposite of animal behavior, and represents a serious gap in performance.

\subsection{Combining curiosity and prioritized replay}
Curious Replay addresses the challenge of model-based adaptation in changing environments. It encourages an accurate, adaptive world model by prioritizing optimization on experiences that have been trained on the fewest times or that are least accurately modeled. We take inspiration from two sources: (1) Prioritized Experience Replay, which uses temporal-difference (TD) error of the value estimate for prioritization \cite{schaul2015prioritized}; and (2) Curiosity as an intrinsic motivation, which has received much attention as a signal for guiding exploration in sparsely-rewarded environments \cite{schmidhuber1991possibility, oudeyer2007intrinsic}. Such curiosity signals include count-based novelty \cite{bellemare2016unifying, tang2017exploration} or adversarial model-error \cite{stadie2015incentivizing, pathak2017curiosity, haber2018learning, guo2022byol}. In the following, we describe versions of these two signals used to guide replay. By combining them into our method, Curious Replay, we improve Dreamer's ability to adapt to changing environments---as demonstrated in Figure \ref{fig2}, for example, where it dramatically enhances adaptation speed and world model accuracy.

\subsection{Count-based Replay}
One hypothesis is that Dreamer's uniform replay buffer sampling does not promote training the world model on new data from the changed environment. Inspired by count-based novelty, we develop Count-based Replay to ensure that the model trains on new data, and thus avoids ignoring potentially valuable data it has collected.

Count-based replay biases sampling towards recent experiences, and ensures that the agent revisits each experience multiple times (with high probability). Prioritization relies on tracking the visit count $v_i$, the number of times an experience has been revisited (i.e. incorporated in a training batch). Here, an experience is a single state transition. The counter $v \in \mathbb{R}^{|R|}$, for buffer capacity $|R|$, is used with hyperparameter $\beta \in [0, 1]$ to prioritize sampling as $p_i = \beta^{v_i}$. The priorities are stored in a SumTree \cite{schaul2015prioritized} to enable efficient normalization to a sampling probability.

\subsection{Adversarial Replay}
Count-based replay is agnostic to the actual content of the experiences. This can be a drawback if particular experiences are challenging to learn, or where certain experiences are redundant with those that have already been learned. We hypothesized that adaptation might be aided by prioritizing experiences that the model is not currently good at. This led to Adversarial Replay, which is inspired by adversarial intrinsic motivation \cite{stadie2015incentivizing, pathak2017curiosity}, and can be similar to the notion of surprise \cite{berseth2019smirl}. In fact, there is evidence of a model-error related signal being helpful for prioritizing replay \cite{oh2021model}, and we tailor this approach to the Dreamer setting. 

Adversarial replay prioritizes experiences that the world model does not accurately predict. It uses model-loss as a prioritization signal, with $p = (|\mathcal{L}| + \epsilon)^\alpha$ for loss $\mathcal{L}$, $\epsilon$ a small positive constant, and $\alpha \in [0, 1]$ that determines the extent of prioritization ($\alpha=0$ is uniform sampling). $\mathcal{L}$ is the loss used to train the world model, which in the case of Dreamer is $\mathcal{L} = \mathcal{L}_{image} + \mathcal{L}_{reward} + \mathcal{L}_{KL}$. For simplicity and speed, the priority for an experience is only updated when it has been trained on. This has the potential to yield stale priorities, but we did not observe any associated deficiencies in our experiments. Priorities are initialized to the maximum value for new experiences added to the replay buffer. One potential consideration is the susceptibility of adversarial curiosity to the white-noise problem --- where completely unpredictable state transitions can prevent the curious agent from being interested in anything else. However, Adversarial Replay is a fundamentally different scenario. Here, the curiosity signal provides prioritization for a fully supervised optimization problem---to fit the world model to the data in the replay buffer---and the experienced state transitions cannot be unpredictable in the same way.

\subsection{Curious Replay (Adversarial + Count)}
To combine benefits of count-based and adversarial replay---prioritizing experiences that are less-frequently replayed or less-understood---we additively merge both approaches into a single Curious Replay prioritization, using a scale factor $c$ to yield an overall priority for experience $i$ of:

\begin{equation}
\centering
p_i = c\beta^{v_i} + (|\mathcal{L}_i| + \epsilon)^\alpha
\label{p_eqn}
\end{equation}

where $\beta$ determines the falloff of count-based priority, $v$ counts how many times each experience has been replayed, $\mathcal{L}_i$ is the loss for an experience, $\epsilon$ is a small positive constant, and $\alpha$ determines the sharpness of adversarial prioritization.

\subsection{Temporal-Difference Prioritized Experience Replay}
It is well known that prioritizing sampling based on value can be beneficial \cite{moore1993prioritized}. Indeed, standard uses of prioritization leverage the temporal-difference error of the value estimate \cite{schaul2015prioritized, horgan2018distributed, hessel2018rainbow}: 
% \begin{equation}
$\delta_t = r_t + \gamma V(x_{t+1}) - V(x_t)$,
% \end{equation}
for value function $V(x)$  (e.g. critic), and $p_i = (|\delta_i| + \epsilon)^\alpha$. We also investigate use of this prioritization in our setting.

\section{Experiments}
In this section, we investigate the following questions:
\begin{enumerate}[noitemsep,nolistsep]
    \item[Q1.] Does Curious Replay consistently outperform other methods in environments with a step change at a single timepoint?
    \item[Q2.] Does Curious Replay consistently outperform other methods in a more complex, open-world environment requiring continual learning, such as Crafter?
    \item[Q3.] How much does each component of Curious Replay contribute?
    \item[Q4.] How much impact does Curious Replay have on performance in unchanging environments? 
    \item[Q5.] Does a Curious Replay agent retain the ability to choose good actions and make good world predictions for earlier states that it has not recently experienced?
\end{enumerate}

\subsection{Experimental Setup}

\paragraph{Environment Details} 

Visual observations are 64 × 64 × 3 pixel renderings. The Control Suite performance metric is extrinsic reward return. For object interaction, the primary metric is number of environment steps until the 5th object interaction (a metric determined in part by mouse experiments in Figure \ref{figS1}, but results are robust to the exact choice of interaction number, see Figure \ref{fig_Nth}). For Crafter, the score is the geometric mean of achievement success rates, to account for the difficulty of different possible achievements.

\paragraph{Baselines}
Plan2Explore is our baseline for object interaction (DreamerV2 with latent disagreement as intrinsic motivation \cite{sekar2020planning}). We did not find other suitable baseline agents that were model-based, implemented with intrinsic-motivation, and with publicly-available code (e.g. AdA does not have a public implementation  \cite{team2023human}). DreamerV2 \cite{hafner2020mastering} is our baseline for Constrained Control Suite investigations. For Crafter experiments, we include a number of baseline comparisons, including DreamerV2, DreamerPro \cite{deng2022dreamerpro}, IRIS \cite{micheli2022transformers}, and DreamerV3 (\cite{hafner2023mastering}, for which code was not publicly available until after initial submission). In these settings, we also compare against augmenting the baseline agent with temporal-difference (TD) prioritized replay. We additionally compare against DreamerV3 and DreamerV2 on the full Deepmind Control Suite.

\paragraph{Hyperparameters}
Agents used defaults for DreamerV2, DreamerV3, Plan2Explore, and DreamerPro. For Curious Replay, $\beta=0.7$, $\alpha=0.7$, $c=1e4$, $\epsilon=0.01$, and $p_{MAX} = 1e5$. These were optimized on the object interaction assay and fixed across all environments.

\subsection{Object interaction in a changing environment} 

To answer Q1, we use the object interaction assay. 
Curious Replay outperforms Plan2Explore and Plan2Explore w/ TD, interacting within 50K steps, which is more than six times faster than Plan2Explore (Figure \ref{fig2}). TD slightly improves Plan2Explore, but also has high variance. The Curious Replay agent quickly interacts many times with the object (at a faster pace than the other methods), and then starts to level off, yielding a total count by 900k steps that is similar to that of the other two agents (Figure \ref{fig_cumulative}). Figure \ref{fig2} shows the difference in world model performance.  Example agent trajectories demonstrate a clear difference in behavior with Curious Replay (Figure \ref{fig_example_trajectories}). Curious Replay also outperforms TD+Adversarial (Figure \ref{md_comparison}). We also assess Curious Replay in an unchanging environment, where the object is present from the outset. Baseline Plan2Explore performs much better in this setting, but performs no better than Curious Replay (Figure \ref{s1s2}). The conclusions remain the same if different metrics of time-to-object interaction are used (Figure \ref{fig_Nth}). Curious Replay is relatively insensitive to its hyperparameters; all tested values yielded improved object interaction relative to Plan2Explore (Table \ref{hyperparam_sensitivity}).

\subsection{Constrained Control Suite}
 We further answer Q1 with the Constrained Control Suite, and find that on all three tasks, Cup Catch, Cartpole Swingup Sparse, and Cheetah Run, DreamerV2 w/ Curious Replay adapts faster than DreamerV2, (Figure \ref{cdmc}). Quantified at step 750K, Curious Replay has an average return across tasks of 508 versus 121 for DreamerV2.

\begin{figure*}[t]
\vskip 0.2in
\begin{center}
\centerline{\includegraphics[width=\textwidth]{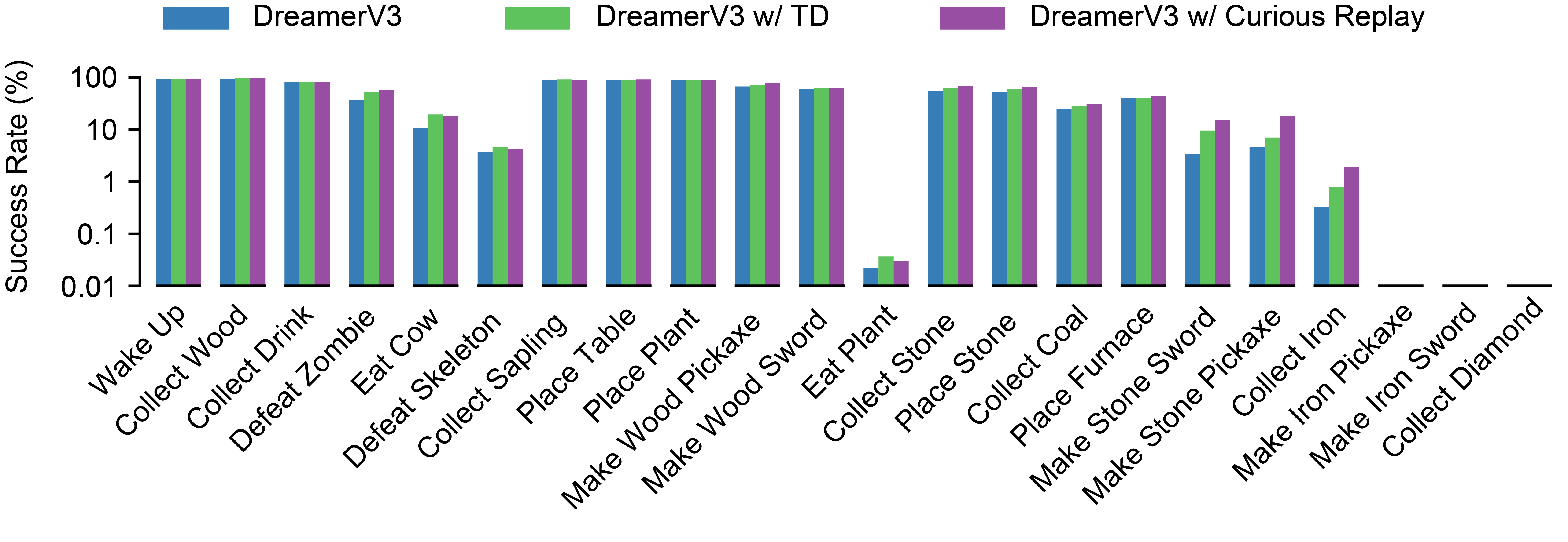}}
\vspace{-15pt}
\caption{Agent ability spectrum for Crafter, ordered left to right by number of prerequisites for an achievement. (n=10 each).}
\vspace{-5pt}
\label{crafter_spectra_dv3}
\end{center}
\vskip -0.3in
\end{figure*}

\subsection{Crafter}
Answering Q2 in Table \ref{crafter_scores_table}, Curious Replay substantially improves performance on Crafter. DreamerV3 w/ Curious Replay achieves a new state-of-the-art score on Crafter. Moreover, for each baseline (DreamerV3, DreamerV2, DreamerPro, and Plan2Explore), Curious Replay has better performance than the baseline as well as the baseline w/ TD prioritization. This high score is reflected in the agent ability spectrum (Figure \ref{crafter_spectra_dv3}), with Curious Replay exhibiting higher success at the more challenging achievements such as Collect Iron and Make Stone Pickaxe. Impressively, there were even a few rare instances when Curious Replay succeeded at an achievement never reached by DreamerV3, with 2/10 seeds of DreamerV3 w/ Curious Replay succeeding at Make Iron Sword  (and 1/10 seeds of DreamerV3 w/ TD succeeding) . 
Curious Replay also improves the ability spectrum of Plan2Explore (Figures \ref{fig_spectrum_unsupervised}, \ref{crafter_score_unsupervised}) and DreamerV2 (Figure \ref{crafter_spectrum}), especially for more challenging achievements. 

We assessed the hyperparameter sensitivity of Curious Replay's performance on Crafter, using DreamerV2 (Table \ref{hyperparam_sensitivity_crafter}). We found that while all tested CR hyperparameters (including  our default hyperparameters $\alpha$=0.7, $\beta$=0.7) yielded a higher mean score than DreamerV2, it was also possible to obtain an even higher score through hyperparameter search.

We also tested whether increasing the train ratio could improve the performance of Curious Replay \cite{d2022sample}. 
Highlighted in Table \ref{train_ratio_table}, we found that increasing the train frequency by 8x improves the Crafter score of DreamerV2 w/ Curious Replay (to 15.7 $\pm$ 2.4) but not DreamerV2.

\begin{table}[hb!]
\vskip -0.3in

\caption{Crafter scores compared to previous algorithms. DreamerV3 with Curious Replay surpasses DreamerV3 to achieve a new state-of-the-art score. (CR = Curious Replay, TD = temporal difference prioritization, $n$=10 seeds per method, mean $\pm$ s.d.). $^*$=\mbox{with optimized CR hyperparameters, see Table \ref{hyperparam_sensitivity_crafter}}.}
\begin{center}

\begin{NiceTabular}{| l | r |}
\toprule

\textbf{Method} & \multicolumn{1}{c}{\textbf{Crafter Score}}  \\
\noalign{\smallskip}\hline\noalign{\smallskip}

DreamerV3 CR & \textbf{19.4 $\pm$ 1.6}\% \\
DreamerV3 TD & 17.0 $\pm$ 2.0\% \\
DreamerV2 CR (8x train freq.) & 15.7 $\pm$ 2.4\% \\
DreamerV3$^\dagger$ & 14.5 $\pm$ 1.6\% \\
DreamerV2 CR$^*$ & 13.2 $\pm$ 1.4\% \\
LSTM-SPCNN$^\dagger$ & 12.1 $\pm$ 0.8\% \\
DreamerV2 & 11.7 $\pm$ 0.5\% \\
DreamerV2 (8x train freq.) & 11.0 $\pm$ 1.5\% \\
DreamerV2 TD & 10.8 $\pm$ 0.6\% \\
DreamerV2$^\dagger$ & 10.0 $\pm$ 1.2\% \\
DreamerPro CR & 6.8 $\pm$ 0.5\% \\
DreamerPro TD & 5.8 $\pm$ 0.5\% \\
DreamerPro & 4.7 $\pm$ 0.5\% \\
IRIS & 4.6 $\pm$ 0.7\% \\
Plan2Explore CR (unsup) & 2.7 $\pm$ 0.1\% \\ %2.71
Plan2Explore TD (unsup) & 2.7  $\pm$ 0.1\% \\ %2.65
Plan2Explore (unsup) & 2.2 $\pm$ 0.1\% \\ %2.15
\bottomrule
\end{NiceTabular}
$^\dagger$\mbox{Published  results, see \cite{hafner2023mastering}}
\end{center}
\label{crafter_scores_table}
\end{table}

\paragraph{Why does CR help on crafter?}
The set of resources held by the Crafter agent can be considered to be a new phase $\chi$ of the POMDP (see Section 2.1), and discovering a new resource means there is a new phase to learn about. We find that Curious Replay prioritizes and trains more frequently on experiences where the agent is holding recently discovered resources. To show this, we focused on the chain of achievements that includes ``collecting iron," which the Curious Replay agent achieves at a higher rate than the baseline. To collect iron, a number of intermediate resources and tools must first be acquired sequentially: wood, a wood pickaxe, stone, a stone pickaxe, and then iron.

Experiences where the agent holds each of these resources beyond wood have a relative sampling probability $>1$, meaning they are more likely to be sampled than with a uniform distribution (Table \ref{crafter_sampling_prob}). Additionally, the experiences of holding recently discovered resources are replayed more often during training (Table \ref{crafter_training_count}). The effect appears magnified for resources further in the chain, where there is a larger increase in relative training frequency.

The behavioral impact of Curious Replay is faster progression through the sequence of achievements. We analyzed the steps required for the average agent to reach 1\% success on an achievement, in a 20K step moving average. Curious Replay leads to a substantially faster progression through the sequence of achievements, ultimately helping the agent succeed at more challenging achievements (Table \ref{achievement_progression}).

\subsection{Ablations}
Answering Q3, we assessed individual contributions of the Adversarial and Count-based components of Curious Replay (Table \ref{ablation}). Each improved performance relative to the baseline, but the combination as Curious Replay was better.

\begin{table}[h]
\caption{Performance of ablated versions of Curious Replay (ours), which combines Adversarial and Count Replay. Plan2Explore is used for object interaction, measured by \# of steps to 5th interaction. DreamerV3 is used for Crafter, measured by the score.
}
\label{ablation_table}
\begin{center}
\begin{small}
\begin{NiceTabular}{ |l| r r  r | }
\toprule
 \multicolumn{1}{c}{\textbf{Assay}} &  \multicolumn{1}{c}{\textbf{Adversarial}} &  \multicolumn{1}{c}{\textbf{Count}} &   \multicolumn{1}{c}{\textbf{CR (ours)}} \\

\midrule
Object $\downarrow$    &  1.8 $\pm$ 0.5 & 2.2 $\pm$ 0.7 &  \textbf{0.5 $\pm$ 0.1} \\
Crafter $\uparrow$  & 17.3 $\pm$ 2.7  &  16.2 $\pm$ 1.3 & \textbf{19.4 $\pm$ 1.6}   \\

\bottomrule
\end{NiceTabular}
\end{small}
\end{center}
\label{ablation}
\vskip -0.1in
\end{table}
\vspace{-10pt}

\subsection{Deepmind Control Suite}
Answering Q4, we assessed Curious Replay's impact  in unchanging environments by using the full Deepmind Control Suite (20 tasks). DreamerV3 w/ CR had a similar mean and median score across all 20 tasks as compared with DreamerV3 (Table \ref{standard_dmc}). DreamerV2 w/ CR was slightly worse than DreamerV2.  Interestingly, CR substantially improved performance on tasks such as Quadruped Run whereas it reduced performance on tasks such as Cartpole Swingup Sparse and Pendulum Swingup (Table \ref{dreamerv3_dmc_tasks}, Figures \ref{dmc_dv2_supp}, \ref{dmc_dv3_supp}). There is thus perhaps a pattern, where CR yields improvement on complex locomotion but has a failure mode with sparsely rewarded swingup agents, suggesting an interesting direction for future investigation. Overall, and especially with DreamerV3, Curious Replay has little or no impact on performance across the full unchanging Deepmind Control Suite.  

We additionally assessed whether Curious Replay is impacted by unpredictable environment noise. We used the Distracting Control Suite Walker Walk task with a noisy background, where at each time step the background is a randomly selected image. This experiment tests whether Curious Replay is inordinately hampered by complex and unpredictable distractions. We found that Curious Replay performs just as well as the baseline (Figure \ref{walker_random}).

\begin{table}[h]
\caption{Standard Deepmind Control Suite, mean and median return across 20 tasks, n=3 for DreamerV3 CR, n=2 for DreamerV2 CR. $^\dagger$\mbox{Published  results, see \cite{hafner2023mastering}}. }
\vskip -1in
\begin{center}
\begin{small}
\begin{NiceTabular}{|l|c c|}
\toprule
\textbf{Method} & \textbf{Mean} & \textbf{Median} \\ 
\midrule
DreamerV3 CR   & 734.8 & \textbf{815.1} \\
DreamerV3$^\dagger$    & \textbf{739.6} & 808.5 \\
DreamerV2 CR   & 643.6 & 635.8 \\
DreamerV2    & 715.7 & 770.7 \\

\bottomrule
\end{NiceTabular}
\end{small}
\end{center}
\label{standard_dmc}
\vskip -0.1in
\end{table}

\subsection{Forgetting}
 We address Q5 with two approaches. First, we assess world model performance in a variant of the object interaction assay, where we introduce the object at $T_0=500$K steps and then remove the object at $T_1=1.5$M steps. Using the same approach as in Figure \ref{fig2}, we assess model performance on a set of test episodes. An increase in model error after object removal would signify evidence that the model is forgetting about the object. We find no such evidence for Plan2Explore or Plan2Explore w/ CR (Figure \ref{fig_clearbuffer_error}). To calibrate what catastrophic forgetting might look like, we assessed the performance of Plan2Explore with a replay buffer that is cleared at $T_1$, which has been shown to help with adaptation but cause  catastrophic forgetting \cite{wan2022towards}. Indeed, with a cleared replay buffer the model error dramatically worsens. Thus, Curious Replay is far more resilient to catastrophic forgetting than simply clearing the replay buffer. It achieves this by using a large buffer to store and revisit old experiences, but remains adaptable by using prioritization to focus training on the newer states.

\begin{figure}[h]
\begin{center}
\centerline{\includegraphics[width=\columnwidth]{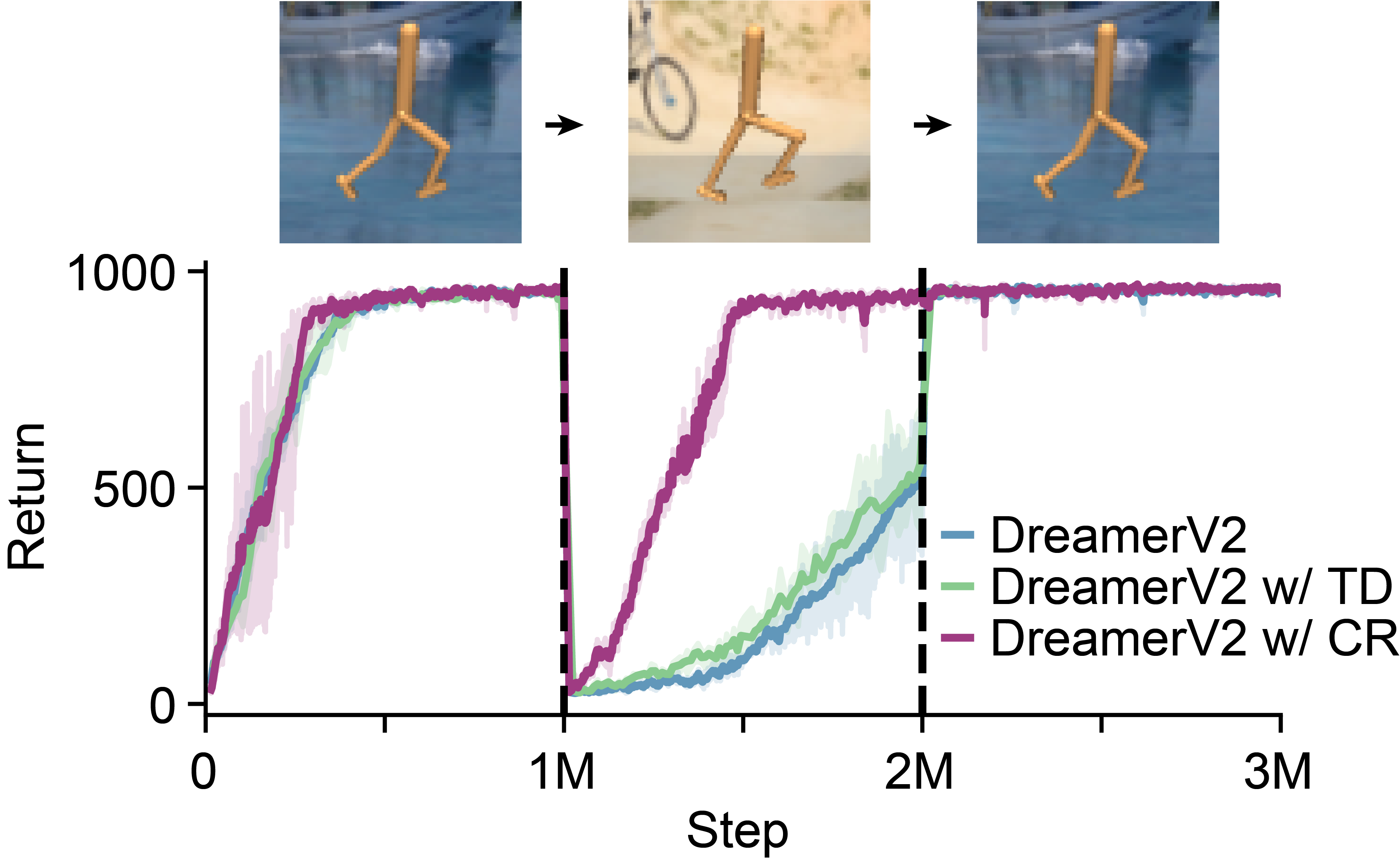}}
\vskip -0.15in
\caption{Background-Swap walker\_walk. Background changes at step 1M, and reverts at step 2M. Curious Replay improves performance after 1M, without degrading performance after 2M. 
}
\label{backgroundswap}
\end{center}
\vskip -0.4in
\end{figure}

 Additionally, we used the Background-Swap Control Suite to assess policy forgetting (Figure \ref{backgroundswap}). Curious Replay yields better adaptation at $T_0=1$M, and matches the baseline after $T_1=2$M, demonstrating that in this experiment Curious Replay does not elicit catastrophic forgetting. This was also true across four additional tasks (Figure \ref{ddmc_supp}, Table \ref{ddmc_table}).

\section{Related work}
\paragraph{Model-based RL} Recent advances have propelled model-based RL to success on classic board games \cite{schrittwieser2020mastering} and on tasks with high dimensional input \cite{ha2018recurrent} including benchmarks such as Deepmind Control Suite \cite{hafner2019dream} and Atari \cite{kaiser2019model, hafner2020mastering}. These successes have been extended in a number of settings \cite{wu2022daydreamer, mendonca2021discovering, hafner2022deep, micheli2022transformers, chen2022transdreamer, deng2022dreamerpro, hafner2023mastering}.

\paragraph*{Adaptation}
Adaptation has been studied as a component of continual learning \cite{thrun1998lifelong, khetarpal2022towards, julian2020never, xie2021deep}. Approaches toward adaptation include using system identification \cite{bongard2006resilient, cully2015robots, kumar2021rma}, metalearning \cite{finn2017model, al2017continuous, song2020rapidly, nagabandi2018deep, yu2020learning}, planning using recent states \cite{wan2022towards}, and changepoint detection \cite{adams2007bayesian, fearnhead2007line, hadoux2014sequential, banerjee2017quickest}.

\paragraph*{Prioritized replay}
Experience replay \cite{lin1992self, mnih2015human} has been enhanced in a number of settings through prioritization. Priority signals include changes in value \cite{moore1993prioritized}, temporal-difference error \cite{schaul2015prioritized, pan2022understanding}, state frequency \cite{novati2019remember, sinha2022experience, sun2020attentive}, and regret \cite{liu2021regret}.
The sampling strategy itself can even be optimized \cite{zha2019experience, oh2020learning, burega2022learning}. 
Pioneering work has also investigated use of model errors for prioritization \cite{oh2021model}, with an implementation and application complementary to ours.

\paragraph*{Intrinsically-motivated RL}
Curiosity-based intrinsic motivation has emerged as a valuable strategy for guiding exploration in sparsely-rewarded environments \cite{schmidhuber1991possibility, oudeyer2007intrinsic}. Such signals can include model error \cite{stadie2015incentivizing, pathak2017curiosity, guo2022byol}, surprise \cite{achiam2017surprise}, model learning progress \cite{kim2020active}, model uncertainty \cite{pathak2019self, sekar2020planning}, latent state novelty \cite{raileanu2020ride},  information gain \cite{still2012information, houthooft2016vime}, state novelty \cite{bellemare2016unifying, machado2020count, burda2018exploration, yarats2021reinforcement, tang2017exploration, parisi2021interesting}, diversity \cite{eysenbach2018diversity}, and empowerment \cite{klyubin2005all, mohamed2015variational, gregor2016variational}.

\section{Discussion}

\paragraph{Limitations}  There may be instances in which the Curious Replay signal is not aligned with a task, potentially because the task requires accuracy on a very specific subset of the world. Future work may incorporate task-related prioritization signals, or use different prioritizations for updating the world model versus the policy. Curious Replay is also not currently tailored to account for changing reward functions, as in the LoCA setting, which will be an interesting direction to investigate \cite{wan2022towards, van2020loca}. Although there was no evidence of catastrophic forgetting in our experiments, further steps can be taken mitigate possible forgetting, such as occasionally recalculating the prioritization weights across the buffer. Additionally, substantially increasing the speed of model training may mitigate the need for replay prioritization by allowing the model to frequently train on the entire replay buffer. Finally, although Curious Replay improved the speed to object interaction, agent behavior still does not fully mirror animal behavior (Figure \ref{figS0}), leaving room for future work. 

\paragraph{Animal inspiration} At its core, Curious Replay was inspired by animal behavior. The prowess of animals at adapting to changing environments  highlighted clear deficiencies in existing model-based agents, and led us to develop Curious Replay to fix these shortcomings. The  effectiveness of Curious Replay may allow the favor to be returned, by providing inspiration to the study of animal physiology. For example, Curious Replay  suggests hypotheses about how animals might replay their past: experiences that are more recent or surprising should be replayed more frequently. Intriguingly, recent experiments have begun to uncover precisely this phenomenon, using electrical recordings of hippocampal replay in rats as they explore environments with varying levels of familiarity \cite{gorriz2023role}.

\paragraph{Outlook} We present Curious Replay, a method that improves agent adaptation in changing environments. Inspired by the use of curiosity as an intrinsic motivation for selecting actions, we use curiosity signals for selecting experiences to use during training. Future directions include incorporating Curious Replay into other model-based and model-free agents, investigating use of additional curiosity signals, and testing performance in an even wider variety of settings. In sum, by using curiosity to guide experience replay, Curious Replay opens avenues for more effective world model training, exploration, and adaptation.

\section{Acknowledgements}
I.K. is a Merck Awardee of the Life Sciences Research Foundation, and a Wu Tsai Stanford Neurosciences Institute Interdisciplinary Scholar. Thank you to Karl Deisseroth for access to mice for animal experiments. This work is also in part funded by Human-Centered AI Hoffman-Yee and Google Cloud Credit Grants, the Stanford Graduate School of Education and the Stanford Accelerator for Learning.

\clearpage

\newpage
\bibliography{main}

\begin{thebibliography}{110}
\providecommand{\natexlab}[1]{#1}
\providecommand{\url}[1]{\texttt{#1}}
\expandafter\ifx\csname urlstyle\endcsname\relax
  \providecommand{\doi}[1]{doi: #1}\else
  \providecommand{\doi}{doi: \begingroup \urlstyle{rm}\Url}\fi

\bibitem[Abadi et~al.(2015)Abadi, Agarwal, Barham, Brevdo, Chen, Citro,
  Corrado, Davis, Dean, Devin, Ghemawat, Goodfellow, Harp, Irving, Isard, Jia,
  Jozefowicz, Kaiser, Kudlur, Levenberg, Man\'{e}, Monga, Moore, Murray, Olah,
  Schuster, Shlens, Steiner, Sutskever, Talwar, Tucker, Vanhoucke, Vasudevan,
  Vi\'{e}gas, Vinyals, Warden, Wattenberg, Wicke, Yu, and
  Zheng]{tensorflow2015-whitepaper}
Abadi, M., Agarwal, A., Barham, P., Brevdo, E., Chen, Z., Citro, C., Corrado,
  G.~S., Davis, A., Dean, J., Devin, M., Ghemawat, S., Goodfellow, I., Harp,
  A., Irving, G., Isard, M., Jia, Y., Jozefowicz, R., Kaiser, L., Kudlur, M.,
  Levenberg, J., Man\'{e}, D., Monga, R., Moore, S., Murray, D., Olah, C.,
  Schuster, M., Shlens, J., Steiner, B., Sutskever, I., Talwar, K., Tucker, P.,
  Vanhoucke, V., Vasudevan, V., Vi\'{e}gas, F., Vinyals, O., Warden, P.,
  Wattenberg, M., Wicke, M., Yu, Y., and Zheng, X.
\newblock {TensorFlow}: Large-scale machine learning on heterogeneous systems,
  2015.
\newblock URL \url{https://www.tensorflow.org/}.
\newblock Software available from tensorflow.org.

\bibitem[Abbas et~al.(2020)Abbas, Sokota, Talvitie, and White]{abbas20a}
Abbas, Z., Sokota, S., Talvitie, E., and White, M.
\newblock Selective dyna-style planning under limited model capacity.
\newblock In III, H.~D. and Singh, A. (eds.), \emph{Proceedings of the 37th
  International Conference on Machine Learning}, volume 119 of
  \emph{Proceedings of Machine Learning Research}, pp.\  1--10. PMLR, 13--18
  Jul 2020.
\newblock URL \url{https://proceedings.mlr.press/v119/abbas20a.html}.

\bibitem[Achiam \& Sastry(2017)Achiam and Sastry]{achiam2017surprise}
Achiam, J. and Sastry, S.
\newblock Surprise-based intrinsic motivation for deep reinforcement learning.
\newblock \emph{arXiv preprint arXiv:1703.01732}, 2017.

\bibitem[Adams \& MacKay(2007)Adams and MacKay]{adams2007bayesian}
Adams, R.~P. and MacKay, D.~J.
\newblock Bayesian online changepoint detection.
\newblock \emph{arXiv preprint arXiv:0710.3742}, 2007.

\bibitem[Ahmadlou et~al.(2021)Ahmadlou, Houba, van Vierbergen, Giannouli,
  Gimenez, van Weeghel, Darbanfouladi, Shirazi, Dziubek, Kacem,
  et~al.]{ahmadlou2021cell}
Ahmadlou, M., Houba, J.~H., van Vierbergen, J.~F., Giannouli, M., Gimenez,
  G.-A., van Weeghel, C., Darbanfouladi, M., Shirazi, M.~Y., Dziubek, J.,
  Kacem, M., et~al.
\newblock A cell type--specific cortico-subcortical brain circuit for
  investigatory and novelty-seeking behavior.
\newblock \emph{Science}, 372\penalty0 (6543):\penalty0 eabe9681, 2021.

\bibitem[Al-Shedivat et~al.(2017)Al-Shedivat, Bansal, Burda, Sutskever,
  Mordatch, and Abbeel]{al2017continuous}
Al-Shedivat, M., Bansal, T., Burda, Y., Sutskever, I., Mordatch, I., and
  Abbeel, P.
\newblock Continuous adaptation via meta-learning in nonstationary and
  competitive environments.
\newblock \emph{arXiv preprint arXiv:1710.03641}, 2017.

\bibitem[Balloch et~al.(2023)Balloch, Lin, Wright, Peng, Hussain, Srinivas,
  Kim, and Riedl]{balloch2023neuro}
Balloch, J., Lin, Z., Wright, R., Peng, X., Hussain, M., Srinivas, A., Kim, J.,
  and Riedl, M.~O.
\newblock Neuro-symbolic world models for adapting to open world novelty.
\newblock \emph{arXiv preprint arXiv:2301.06294}, 2023.

\bibitem[Banerjee et~al.(2017)Banerjee, Liu, and How]{banerjee2017quickest}
Banerjee, T., Liu, M., and How, J.~P.
\newblock Quickest change detection approach to optimal control in markov
  decision processes with model changes.
\newblock In \emph{2017 American control conference (ACC)}, pp.\  399--405.
  IEEE, 2017.

\bibitem[Bellemare et~al.(2016)Bellemare, Srinivasan, Ostrovski, Schaul,
  Saxton, and Munos]{bellemare2016unifying}
Bellemare, M., Srinivasan, S., Ostrovski, G., Schaul, T., Saxton, D., and
  Munos, R.
\newblock Unifying count-based exploration and intrinsic motivation.
\newblock \emph{Advances in neural information processing systems}, 29, 2016.

\bibitem[Berseth et~al.(2019)Berseth, Geng, Devin, Rhinehart, Finn, Jayaraman,
  and Levine]{berseth2019smirl}
Berseth, G., Geng, D., Devin, C., Rhinehart, N., Finn, C., Jayaraman, D., and
  Levine, S.
\newblock Smirl: Surprise minimizing reinforcement learning in unstable
  environments.
\newblock \emph{arXiv preprint arXiv:1912.05510}, 2019.

\bibitem[Bodnar et~al.(2020)Bodnar, Hausman, Dulac-Arnold, and
  Jonschkowski]{bodnar2020geometric}
Bodnar, C., Hausman, K., Dulac-Arnold, G., and Jonschkowski, R.
\newblock A geometric perspective on self-supervised policy adaptation.
\newblock \emph{arXiv preprint arXiv:2011.07318}, 2020.

\bibitem[Bongard et~al.(2006)Bongard, Zykov, and Lipson]{bongard2006resilient}
Bongard, J., Zykov, V., and Lipson, H.
\newblock Resilient machines through continuous self-modeling.
\newblock \emph{Science}, 314\penalty0 (5802):\penalty0 1118--1121, 2006.

\bibitem[Burda et~al.(2018)Burda, Edwards, Storkey, and
  Klimov]{burda2018exploration}
Burda, Y., Edwards, H., Storkey, A., and Klimov, O.
\newblock Exploration by random network distillation.
\newblock \emph{arXiv preprint arXiv:1810.12894}, 2018.

\bibitem[Burega et~al.(2022)Burega, Martin, and Bowling]{burega2022learning}
Burega, B., Martin, J.~D., and Bowling, M.
\newblock Learning to prioritize planning updates in model-based reinforcement
  learning.
\newblock In \emph{Sixth Workshop on Meta-Learning at the Conference on Neural
  Information Processing Systems}, 2022.

\bibitem[Buzzega et~al.(2021)Buzzega, Boschini, Porrello, and
  Calderara]{buzzega2021rethinking}
Buzzega, P., Boschini, M., Porrello, A., and Calderara, S.
\newblock Rethinking experience replay: a bag of tricks for continual learning.
\newblock In \emph{2020 25th International Conference on Pattern Recognition
  (ICPR)}, pp.\  2180--2187. IEEE, 2021.

\bibitem[Cassirer et~al.(2021)Cassirer, Barth-Maron, Brevdo, Ramos, Boyd,
  Sottiaux, and Kroiss]{cassirer2021reverb}
Cassirer, A., Barth-Maron, G., Brevdo, E., Ramos, S., Boyd, T., Sottiaux, T.,
  and Kroiss, M.
\newblock Reverb: A framework for experience replay, 2021.

\bibitem[Chen et~al.(2022)Chen, Wu, Yoon, and Ahn]{chen2022transdreamer}
Chen, C., Wu, Y.-F., Yoon, J., and Ahn, S.
\newblock Transdreamer: Reinforcement learning with transformer world models.
\newblock \emph{arXiv preprint arXiv:2202.09481}, 2022.

\bibitem[Clark(2015)]{clark2015pillow}
Clark, A.
\newblock Pillow (pil fork) documentation, 2015.
\newblock URL
  \url{https://buildmedia.readthedocs.org/media/pdf/pillow/latest/pillow.pdf}.

\bibitem[Cully et~al.(2015)Cully, Clune, Tarapore, and Mouret]{cully2015robots}
Cully, A., Clune, J., Tarapore, D., and Mouret, J.-B.
\newblock Robots that can adapt like animals.
\newblock \emph{Nature}, 521\penalty0 (7553):\penalty0 503--507, 2015.

\bibitem[Da~Silva et~al.(2006)Da~Silva, Basso, Bazzan, and
  Engel]{da2006dealing}
Da~Silva, B.~C., Basso, E.~W., Bazzan, A.~L., and Engel, P.~M.
\newblock Dealing with non-stationary environments using context detection.
\newblock In \emph{Proceedings of the 23rd international conference on Machine
  learning}, pp.\  217--224, 2006.

\bibitem[Deng et~al.(2022)Deng, Jang, and Ahn]{deng2022dreamerpro}
Deng, F., Jang, I., and Ahn, S.
\newblock Dreamerpro: Reconstruction-free model-based reinforcement learning
  with prototypical representations.
\newblock In \emph{International Conference on Machine Learning}, pp.\
  4956--4975. PMLR, 2022.

\bibitem[D'Oro et~al.(2022)D'Oro, Schwarzer, Nikishin, Bacon, Bellemare, and
  Courville]{d2022sample}
D'Oro, P., Schwarzer, M., Nikishin, E., Bacon, P.-L., Bellemare, M.~G., and
  Courville, A.
\newblock Sample-efficient reinforcement learning by breaking the replay ratio
  barrier.
\newblock In \emph{Deep Reinforcement Learning Workshop NeurIPS}, 2022.

\bibitem[Eysenbach et~al.(2018)Eysenbach, Gupta, Ibarz, and
  Levine]{eysenbach2018diversity}
Eysenbach, B., Gupta, A., Ibarz, J., and Levine, S.
\newblock Diversity is all you need: Learning skills without a reward function.
\newblock \emph{arXiv preprint arXiv:1802.06070}, 2018.

\bibitem[Fearnhead \& Liu(2007)Fearnhead and Liu]{fearnhead2007line}
Fearnhead, P. and Liu, Z.
\newblock On-line inference for multiple changepoint problems.
\newblock \emph{Journal of the Royal Statistical Society: Series B (Statistical
  Methodology)}, 69\penalty0 (4):\penalty0 589--605, 2007.

\bibitem[Finn et~al.(2017)Finn, Abbeel, and Levine]{finn2017model}
Finn, C., Abbeel, P., and Levine, S.
\newblock Model-agnostic meta-learning for fast adaptation of deep networks.
\newblock In \emph{International conference on machine learning}, pp.\
  1126--1135. PMLR, 2017.

\bibitem[Finn et~al.(2019)Finn, Rajeswaran, Kakade, and Levine]{finn2019online}
Finn, C., Rajeswaran, A., Kakade, S., and Levine, S.
\newblock Online meta-learning.
\newblock In \emph{International Conference on Machine Learning}, pp.\
  1920--1930. PMLR, 2019.

\bibitem[Glickman \& Sroges(1966)Glickman and Sroges]{glickman1966curiosity}
Glickman, S.~E. and Sroges, R.~W.
\newblock Curiosity in zoo animals.
\newblock \emph{Behaviour}, 26\penalty0 (1-2):\penalty0 151--187, 1966.

\bibitem[Gorriz et~al.(2023)Gorriz, Takigawa, and Bendor]{gorriz2023role}
Gorriz, M.~H., Takigawa, M., and Bendor, D.
\newblock The role of experience in prioritizing hippocampal replay.
\newblock \emph{bioRxiv}, pp.\  2023--03, 2023.

\bibitem[Gregor et~al.(2016)Gregor, Rezende, and
  Wierstra]{gregor2016variational}
Gregor, K., Rezende, D.~J., and Wierstra, D.
\newblock Variational intrinsic control.
\newblock \emph{arXiv preprint arXiv:1611.07507}, 2016.

\bibitem[Guo et~al.(2022)Guo, Thakoor, P{\^\i}slar, Pires, Altch{\'e}, Tallec,
  Saade, Calandriello, Grill, Tang, et~al.]{guo2022byol}
Guo, Z.~D., Thakoor, S., P{\^\i}slar, M., Pires, B.~A., Altch{\'e}, F., Tallec,
  C., Saade, A., Calandriello, D., Grill, J.-B., Tang, Y., et~al.
\newblock Byol-explore: Exploration by bootstrapped prediction.
\newblock \emph{arXiv preprint arXiv:2206.08332}, 2022.

\bibitem[Ha \& Schmidhuber(2018)Ha and Schmidhuber]{ha2018recurrent}
Ha, D. and Schmidhuber, J.
\newblock Recurrent world models facilitate policy evolution.
\newblock \emph{Advances in neural information processing systems}, 31, 2018.

\bibitem[Haber et~al.(2018)Haber, Mrowca, Wang, Fei-Fei, and
  Yamins]{haber2018learning}
Haber, N., Mrowca, D., Wang, S., Fei-Fei, L.~F., and Yamins, D.~L.
\newblock Learning to play with intrinsically-motivated, self-aware agents.
\newblock \emph{Advances in neural information processing systems}, 31, 2018.

\bibitem[Hadoux et~al.(2014)Hadoux, Beynier, and Weng]{hadoux2014sequential}
Hadoux, E., Beynier, A., and Weng, P.
\newblock Sequential decision-making under non-stationary environments via
  sequential change-point detection.
\newblock In \emph{Learning over multiple contexts (LMCE)}, 2014.

\bibitem[Hafner(2021)]{hafner2021crafter}
Hafner, D.
\newblock Benchmarking the spectrum of agent capabilities.
\newblock \emph{arXiv preprint arXiv:2109.06780}, 2021.

\bibitem[Hafner et~al.(2019{\natexlab{a}})Hafner, Lillicrap, Ba, and
  Norouzi]{hafner2019dream}
Hafner, D., Lillicrap, T., Ba, J., and Norouzi, M.
\newblock Dream to control: Learning behaviors by latent imagination.
\newblock \emph{arXiv preprint arXiv:1912.01603}, 2019{\natexlab{a}}.

\bibitem[Hafner et~al.(2019{\natexlab{b}})Hafner, Lillicrap, Fischer, Villegas,
  Ha, Lee, and Davidson]{hafner2019learning}
Hafner, D., Lillicrap, T., Fischer, I., Villegas, R., Ha, D., Lee, H., and
  Davidson, J.
\newblock Learning latent dynamics for planning from pixels.
\newblock In \emph{International conference on machine learning}, pp.\
  2555--2565. PMLR, 2019{\natexlab{b}}.

\bibitem[Hafner et~al.(2020)Hafner, Lillicrap, Norouzi, and
  Ba]{hafner2020mastering}
Hafner, D., Lillicrap, T., Norouzi, M., and Ba, J.
\newblock Mastering atari with discrete world models.
\newblock \emph{arXiv preprint arXiv:2010.02193}, 2020.

\bibitem[Hafner et~al.(2022)Hafner, Lee, Fischer, and Abbeel]{hafner2022deep}
Hafner, D., Lee, K.-H., Fischer, I., and Abbeel, P.
\newblock Deep hierarchical planning from pixels.
\newblock \emph{arXiv preprint arXiv:2206.04114}, 2022.

\bibitem[Hafner et~al.(2023)Hafner, Pasukonis, Ba, and
  Lillicrap]{hafner2023mastering}
Hafner, D., Pasukonis, J., Ba, J., and Lillicrap, T.
\newblock Mastering diverse domains through world models.
\newblock \emph{arXiv preprint arXiv:2301.04104}, 2023.

\bibitem[Hansen et~al.(2020)Hansen, Jangir, Sun, Aleny{\`a}, Abbeel, Efros,
  Pinto, and Wang]{hansen2020self}
Hansen, N., Jangir, R., Sun, Y., Aleny{\`a}, G., Abbeel, P., Efros, A.~A.,
  Pinto, L., and Wang, X.
\newblock Self-supervised policy adaptation during deployment.
\newblock \emph{arXiv preprint arXiv:2007.04309}, 2020.

\bibitem[Harris et~al.(2020)Harris, Millman, van~der Walt, Gommers, Virtanen,
  Cournapeau, Wieser, Taylor, Berg, Smith, Kern, Picus, Hoyer, van Kerkwijk,
  Brett, Haldane, del R{\'{i}}o, Wiebe, Peterson, G{\'{e}}rard-Marchant,
  Sheppard, Reddy, Weckesser, Abbasi, Gohlke, and Oliphant]{harris2020array}
Harris, C.~R., Millman, K.~J., van~der Walt, S.~J., Gommers, R., Virtanen, P.,
  Cournapeau, D., Wieser, E., Taylor, J., Berg, S., Smith, N.~J., Kern, R.,
  Picus, M., Hoyer, S., van Kerkwijk, M.~H., Brett, M., Haldane, A., del
  R{\'{i}}o, J.~F., Wiebe, M., Peterson, P., G{\'{e}}rard-Marchant, P.,
  Sheppard, K., Reddy, T., Weckesser, W., Abbasi, H., Gohlke, C., and Oliphant,
  T.~E.
\newblock Array programming with {NumPy}.
\newblock \emph{Nature}, 585\penalty0 (7825):\penalty0 357--362, September
  2020.
\newblock \doi{10.1038/s41586-020-2649-2}.
\newblock URL \url{https://doi.org/10.1038/s41586-020-2649-2}.

\bibitem[Harrison et~al.(2020)Harrison, Sharma, Finn, and
  Pavone]{harrison2020continuous}
Harrison, J., Sharma, A., Finn, C., and Pavone, M.
\newblock Continuous meta-learning without tasks.
\newblock \emph{Advances in neural information processing systems},
  33:\penalty0 17571--17581, 2020.

\bibitem[Hessel et~al.(2018)Hessel, Modayil, Van~Hasselt, Schaul, Ostrovski,
  Dabney, Horgan, Piot, Azar, and Silver]{hessel2018rainbow}
Hessel, M., Modayil, J., Van~Hasselt, H., Schaul, T., Ostrovski, G., Dabney,
  W., Horgan, D., Piot, B., Azar, M., and Silver, D.
\newblock Rainbow: Combining improvements in deep reinforcement learning.
\newblock In \emph{Thirty-second AAAI conference on artificial intelligence},
  2018.

\bibitem[Horgan et~al.(2018)Horgan, Quan, Budden, Barth-Maron, Hessel,
  Van~Hasselt, and Silver]{horgan2018distributed}
Horgan, D., Quan, J., Budden, D., Barth-Maron, G., Hessel, M., Van~Hasselt, H.,
  and Silver, D.
\newblock Distributed prioritized experience replay.
\newblock \emph{arXiv preprint arXiv:1803.00933}, 2018.

\bibitem[Houthooft et~al.(2016)Houthooft, Chen, Duan, Schulman, De~Turck, and
  Abbeel]{houthooft2016vime}
Houthooft, R., Chen, X., Duan, Y., Schulman, J., De~Turck, F., and Abbeel, P.
\newblock Vime: Variational information maximizing exploration.
\newblock \emph{Advances in neural information processing systems}, 29, 2016.

\bibitem[Hunter(2007)]{Hunter:2007}
Hunter, J.~D.
\newblock Matplotlib: A 2d graphics environment.
\newblock \emph{Computing in Science \& Engineering}, 9\penalty0 (3):\penalty0
  90--95, 2007.
\newblock \doi{10.1109/MCSE.2007.55}.

\bibitem[Javed \& White(2019)Javed and White]{javed2019meta}
Javed, K. and White, M.
\newblock Meta-learning representations for continual learning.
\newblock \emph{Advances in Neural Information Processing Systems}, 32, 2019.

\bibitem[Julian et~al.(2020)Julian, Swanson, Sukhatme, Levine, Finn, and
  Hausman]{julian2020never}
Julian, R., Swanson, B., Sukhatme, G.~S., Levine, S., Finn, C., and Hausman, K.
\newblock Never stop learning: The effectiveness of fine-tuning in robotic
  reinforcement learning.
\newblock \emph{arXiv preprint arXiv:2004.10190}, 2020.

\bibitem[Kaiser et~al.(2019)Kaiser, Babaeizadeh, Milos, Osinski, Campbell,
  Czechowski, Erhan, Finn, Kozakowski, Levine, et~al.]{kaiser2019model}
Kaiser, L., Babaeizadeh, M., Milos, P., Osinski, B., Campbell, R.~H.,
  Czechowski, K., Erhan, D., Finn, C., Kozakowski, P., Levine, S., et~al.
\newblock Model-based reinforcement learning for atari.
\newblock \emph{arXiv preprint arXiv:1903.00374}, 2019.

\bibitem[Khetarpal et~al.(2022)Khetarpal, Riemer, Rish, and
  Precup]{khetarpal2022towards}
Khetarpal, K., Riemer, M., Rish, I., and Precup, D.
\newblock Towards continual reinforcement learning: A review and perspectives.
\newblock \emph{Journal of Artificial Intelligence Research}, 75:\penalty0
  1401--1476, 2022.

\bibitem[Kim et~al.(2020)Kim, Sano, De~Freitas, Haber, and
  Yamins]{kim2020active}
Kim, K., Sano, M., De~Freitas, J., Haber, N., and Yamins, D.
\newblock Active world model learning with progress curiosity.
\newblock In \emph{International conference on machine learning}, pp.\
  5306--5315. PMLR, 2020.

\bibitem[Kirkpatrick et~al.(2017)Kirkpatrick, Pascanu, Rabinowitz, Veness,
  Desjardins, Rusu, Milan, Quan, Ramalho, Grabska-Barwinska,
  et~al.]{kirkpatrick2017overcoming}
Kirkpatrick, J., Pascanu, R., Rabinowitz, N., Veness, J., Desjardins, G., Rusu,
  A.~A., Milan, K., Quan, J., Ramalho, T., Grabska-Barwinska, A., et~al.
\newblock Overcoming catastrophic forgetting in neural networks.
\newblock \emph{Proceedings of the national academy of sciences}, 114\penalty0
  (13):\penalty0 3521--3526, 2017.

\bibitem[Klyubin et~al.(2005)Klyubin, Polani, and Nehaniv]{klyubin2005all}
Klyubin, A.~S., Polani, D., and Nehaniv, C.~L.
\newblock All else being equal be empowered.
\newblock In \emph{European Conference on Artificial Life}, pp.\  744--753.
  Springer, 2005.

\bibitem[Koh et~al.(2021)Koh, Sagawa, Marklund, Xie, Zhang, Balsubramani, Hu,
  Yasunaga, Phillips, Gao, et~al.]{koh2021wilds}
Koh, P.~W., Sagawa, S., Marklund, H., Xie, S.~M., Zhang, M., Balsubramani, A.,
  Hu, W., Yasunaga, M., Phillips, R.~L., Gao, I., et~al.
\newblock Wilds: A benchmark of in-the-wild distribution shifts.
\newblock In \emph{International Conference on Machine Learning}, pp.\
  5637--5664. PMLR, 2021.

\bibitem[Kumar et~al.(2020)Kumar, Gupta, and Levine]{kumar2020discor}
Kumar, A., Gupta, A., and Levine, S.
\newblock Discor: Corrective feedback in reinforcement learning via
  distribution correction.
\newblock \emph{Advances in Neural Information Processing Systems},
  33:\penalty0 18560--18572, 2020.

\bibitem[Kumar et~al.(2021)Kumar, Fu, Pathak, and Malik]{kumar2021rma}
Kumar, A., Fu, Z., Pathak, D., and Malik, J.
\newblock Rma: Rapid motor adaptation for legged robots.
\newblock \emph{arXiv preprint arXiv:2107.04034}, 2021.

\bibitem[Laskin et~al.(2021)Laskin, Yarats, Liu, Lee, Zhan, Lu, Cang, Pinto,
  and Abbeel]{laskin2021urlb}
Laskin, M., Yarats, D., Liu, H., Lee, K., Zhan, A., Lu, K., Cang, C., Pinto,
  L., and Abbeel, P.
\newblock Urlb: Unsupervised reinforcement learning benchmark.
\newblock \emph{arXiv preprint arXiv:2110.15191}, 2021.

\bibitem[Li \& Hoiem(2017)Li and Hoiem]{li2017learning}
Li, Z. and Hoiem, D.
\newblock Learning without forgetting.
\newblock \emph{IEEE transactions on pattern analysis and machine
  intelligence}, 40\penalty0 (12):\penalty0 2935--2947, 2017.

\bibitem[Lin(1992)]{lin1992self}
Lin, L.-J.
\newblock Self-improving reactive agents based on reinforcement learning,
  planning and teaching.
\newblock \emph{Machine learning}, 8\penalty0 (3):\penalty0 293--321, 1992.

\bibitem[Liu et~al.(2021)Liu, Xue, Pang, Jiang, Xu, and Yu]{liu2021regret}
Liu, X.-H., Xue, Z., Pang, J., Jiang, S., Xu, F., and Yu, Y.
\newblock Regret minimization experience replay in off-policy reinforcement
  learning.
\newblock \emph{Advances in Neural Information Processing Systems},
  34:\penalty0 17604--17615, 2021.

\bibitem[Machado et~al.(2020)Machado, Bellemare, and Bowling]{machado2020count}
Machado, M.~C., Bellemare, M.~G., and Bowling, M.
\newblock Count-based exploration with the successor representation.
\newblock In \emph{Proceedings of the AAAI Conference on Artificial
  Intelligence}, volume~34, pp.\  5125--5133, 2020.

\bibitem[Mendonca et~al.(2021)Mendonca, Rybkin, Daniilidis, Hafner, and
  Pathak]{mendonca2021discovering}
Mendonca, R., Rybkin, O., Daniilidis, K., Hafner, D., and Pathak, D.
\newblock Discovering and achieving goals via world models.
\newblock \emph{Advances in Neural Information Processing Systems},
  34:\penalty0 24379--24391, 2021.

\bibitem[Micheli et~al.(2022)Micheli, Alonso, and
  Fleuret]{micheli2022transformers}
Micheli, V., Alonso, E., and Fleuret, F.
\newblock Transformers are sample efficient world models.
\newblock \emph{arXiv preprint arXiv:2209.00588}, 2022.

\bibitem[Mnih et~al.(2015)Mnih, Kavukcuoglu, Silver, Rusu, Veness, Bellemare,
  Graves, Riedmiller, Fidjeland, Ostrovski, et~al.]{mnih2015human}
Mnih, V., Kavukcuoglu, K., Silver, D., Rusu, A.~A., Veness, J., Bellemare,
  M.~G., Graves, A., Riedmiller, M., Fidjeland, A.~K., Ostrovski, G., et~al.
\newblock Human-level control through deep reinforcement learning.
\newblock \emph{nature}, 518\penalty0 (7540):\penalty0 529--533, 2015.

\bibitem[Moerland et~al.(2023)Moerland, Broekens, Plaat, Jonker,
  et~al.]{moerland2023model}
Moerland, T.~M., Broekens, J., Plaat, A., Jonker, C.~M., et~al.
\newblock Model-based reinforcement learning: A survey.
\newblock \emph{Foundations and Trends{\textregistered} in Machine Learning},
  16\penalty0 (1):\penalty0 1--118, 2023.

\bibitem[Mohamed \& Jimenez~Rezende(2015)Mohamed and
  Jimenez~Rezende]{mohamed2015variational}
Mohamed, S. and Jimenez~Rezende, D.
\newblock Variational information maximisation for intrinsically motivated
  reinforcement learning.
\newblock \emph{Advances in neural information processing systems}, 28, 2015.

\bibitem[Moore \& Atkeson(1993)Moore and Atkeson]{moore1993prioritized}
Moore, A.~W. and Atkeson, C.~G.
\newblock Prioritized sweeping: Reinforcement learning with less data and less
  time.
\newblock \emph{Machine learning}, 13\penalty0 (1):\penalty0 103--130, 1993.

\bibitem[Nagabandi et~al.(2018)Nagabandi, Finn, and Levine]{nagabandi2018deep}
Nagabandi, A., Finn, C., and Levine, S.
\newblock Deep online learning via meta-learning: Continual adaptation for
  model-based rl.
\newblock \emph{arXiv preprint arXiv:1812.07671}, 2018.

\bibitem[Novati \& Koumoutsakos(2019)Novati and
  Koumoutsakos]{novati2019remember}
Novati, G. and Koumoutsakos, P.
\newblock Remember and forget for experience replay.
\newblock In \emph{International Conference on Machine Learning}, pp.\
  4851--4860. PMLR, 2019.

\bibitem[Oh et~al.(2020)Oh, Lee, Shin, Yang, and Hwang]{oh2020learning}
Oh, Y., Lee, K., Shin, J., Yang, E., and Hwang, S.~J.
\newblock Learning to sample with local and global contexts in experience
  replay buffer.
\newblock \emph{arXiv preprint arXiv:2007.07358}, 2020.

\bibitem[Oh et~al.(2021)Oh, Shin, Yang, and Hwang]{oh2021model}
Oh, Y., Shin, J., Yang, E., and Hwang, S.~J.
\newblock Model-augmented prioritized experience replay.
\newblock In \emph{International Conference on Learning Representations}, 2021.

\bibitem[Oudeyer et~al.(2007)Oudeyer, Kaplan, and Hafner]{oudeyer2007intrinsic}
Oudeyer, P.-Y., Kaplan, F., and Hafner, V.~V.
\newblock Intrinsic motivation systems for autonomous mental development.
\newblock \emph{IEEE transactions on evolutionary computation}, 11\penalty0
  (2):\penalty0 265--286, 2007.

\bibitem[Pan et~al.(2020)Pan, Mei, and Farahmand]{pan2020frequency}
Pan, Y., Mei, J., and Farahmand, A.-m.
\newblock Frequency-based search-control in dyna.
\newblock \emph{arXiv preprint arXiv:2002.05822}, 2020.

\bibitem[Pan et~al.(2022)Pan, Mei, Farahmand, White, Yao, Rohani, and
  Luo]{pan2022understanding}
Pan, Y., Mei, J., Farahmand, A.-m., White, M., Yao, H., Rohani, M., and Luo, J.
\newblock Understanding and mitigating the limitations of prioritized replay.
\newblock In \emph{The 38th Conference on Uncertainty in Artificial
  Intelligence}, 2022.

\bibitem[Parisi et~al.(2021)Parisi, Dean, Pathak, and
  Gupta]{parisi2021interesting}
Parisi, S., Dean, V., Pathak, D., and Gupta, A.
\newblock Interesting object, curious agent: Learning task-agnostic
  exploration.
\newblock \emph{Advances in Neural Information Processing Systems},
  34:\penalty0 20516--20530, 2021.

\bibitem[Pathak et~al.(2017)Pathak, Agrawal, Efros, and
  Darrell]{pathak2017curiosity}
Pathak, D., Agrawal, P., Efros, A.~A., and Darrell, T.
\newblock Curiosity-driven exploration by self-supervised prediction.
\newblock In \emph{International conference on machine learning}, pp.\
  2778--2787. PMLR, 2017.

\bibitem[Pathak et~al.(2019)Pathak, Gandhi, and Gupta]{pathak2019self}
Pathak, D., Gandhi, D., and Gupta, A.
\newblock Self-supervised exploration via disagreement.
\newblock In \emph{International conference on machine learning}, pp.\
  5062--5071. PMLR, 2019.

\bibitem[Raileanu \& Rockt{\"a}schel(2020)Raileanu and
  Rockt{\"a}schel]{raileanu2020ride}
Raileanu, R. and Rockt{\"a}schel, T.
\newblock Ride: Rewarding impact-driven exploration for procedurally-generated
  environments.
\newblock \emph{arXiv preprint arXiv:2002.12292}, 2020.

\bibitem[Ritter et~al.(2018)Ritter, Wang, Kurth-Nelson, Jayakumar, Blundell,
  Pascanu, and Botvinick]{ritter2018been}
Ritter, S., Wang, J., Kurth-Nelson, Z., Jayakumar, S., Blundell, C., Pascanu,
  R., and Botvinick, M.
\newblock Been there, done that: Meta-learning with episodic recall.
\newblock In \emph{International conference on machine learning}, pp.\
  4354--4363. PMLR, 2018.

\bibitem[Rolnick et~al.(2019)Rolnick, Ahuja, Schwarz, Lillicrap, and
  Wayne]{rolnick2019experience}
Rolnick, D., Ahuja, A., Schwarz, J., Lillicrap, T., and Wayne, G.
\newblock Experience replay for continual learning.
\newblock \emph{Advances in Neural Information Processing Systems}, 32, 2019.

\bibitem[Rusu et~al.(2016)Rusu, Rabinowitz, Desjardins, Soyer, Kirkpatrick,
  Kavukcuoglu, Pascanu, and Hadsell]{rusu2016progressive}
Rusu, A.~A., Rabinowitz, N.~C., Desjardins, G., Soyer, H., Kirkpatrick, J.,
  Kavukcuoglu, K., Pascanu, R., and Hadsell, R.
\newblock Progressive neural networks.
\newblock \emph{arXiv preprint arXiv:1606.04671}, 2016.

\bibitem[Sagawa et~al.(2021)Sagawa, Koh, Lee, Gao, Xie, Shen, Kumar, Hu,
  Yasunaga, Marklund, et~al.]{sagawa2021extending}
Sagawa, S., Koh, P.~W., Lee, T., Gao, I., Xie, S.~M., Shen, K., Kumar, A., Hu,
  W., Yasunaga, M., Marklund, H., et~al.
\newblock Extending the wilds benchmark for unsupervised adaptation.
\newblock \emph{arXiv preprint arXiv:2112.05090}, 2021.

\bibitem[Schaul et~al.(2015)Schaul, Quan, Antonoglou, and
  Silver]{schaul2015prioritized}
Schaul, T., Quan, J., Antonoglou, I., and Silver, D.
\newblock Prioritized experience replay.
\newblock \emph{arXiv preprint arXiv:1511.05952}, 2015.

\bibitem[Schmidhuber(1991)]{schmidhuber1991possibility}
Schmidhuber, J.
\newblock A possibility for implementing curiosity and boredom in
  model-building neural controllers.
\newblock In \emph{Proc. of the international conference on simulation of
  adaptive behavior: From animals to animats}, pp.\  222--227, 1991.

\bibitem[Schrittwieser et~al.(2020)Schrittwieser, Antonoglou, Hubert, Simonyan,
  Sifre, Schmitt, Guez, Lockhart, Hassabis, Graepel,
  et~al.]{schrittwieser2020mastering}
Schrittwieser, J., Antonoglou, I., Hubert, T., Simonyan, K., Sifre, L.,
  Schmitt, S., Guez, A., Lockhart, E., Hassabis, D., Graepel, T., et~al.
\newblock Mastering atari, go, chess and shogi by planning with a learned
  model.
\newblock \emph{Nature}, 588\penalty0 (7839):\penalty0 604--609, 2020.

\bibitem[Sekar et~al.(2020)Sekar, Rybkin, Daniilidis, Abbeel, Hafner, and
  Pathak]{sekar2020planning}
Sekar, R., Rybkin, O., Daniilidis, K., Abbeel, P., Hafner, D., and Pathak, D.
\newblock Planning to explore via self-supervised world models.
\newblock In \emph{International Conference on Machine Learning}, pp.\
  8583--8592. PMLR, 2020.

\bibitem[Sinha et~al.(2022)Sinha, Song, Garg, and Ermon]{sinha2022experience}
Sinha, S., Song, J., Garg, A., and Ermon, S.
\newblock Experience replay with likelihood-free importance weights.
\newblock In \emph{Learning for Dynamics and Control Conference}, pp.\
  110--123. PMLR, 2022.

\bibitem[Song et~al.(2020)Song, Yang, Choromanski, Caluwaerts, Gao, Finn, and
  Tan]{song2020rapidly}
Song, X., Yang, Y., Choromanski, K., Caluwaerts, K., Gao, W., Finn, C., and
  Tan, J.
\newblock Rapidly adaptable legged robots via evolutionary meta-learning.
\newblock In \emph{2020 IEEE/RSJ International Conference on Intelligent Robots
  and Systems (IROS)}, pp.\  3769--3776. IEEE, 2020.

\bibitem[Stadie et~al.(2015)Stadie, Levine, and
  Abbeel]{stadie2015incentivizing}
Stadie, B.~C., Levine, S., and Abbeel, P.
\newblock Incentivizing exploration in reinforcement learning with deep
  predictive models.
\newblock \emph{arXiv preprint arXiv:1507.00814}, 2015.

\bibitem[Still \& Precup(2012)Still and Precup]{still2012information}
Still, S. and Precup, D.
\newblock An information-theoretic approach to curiosity-driven reinforcement
  learning.
\newblock \emph{Theory in Biosciences}, 131\penalty0 (3):\penalty0 139--148,
  2012.

\bibitem[Stone et~al.(2021)Stone, Ramirez, Konolige, and
  Jonschkowski]{stone2021distracting}
Stone, A., Ramirez, O., Konolige, K., and Jonschkowski, R.
\newblock The distracting control suite--a challenging benchmark for
  reinforcement learning from pixels.
\newblock \emph{arXiv preprint arXiv:2101.02722}, 2021.

\bibitem[Sun et~al.(2020)Sun, Zhou, and Li]{sun2020attentive}
Sun, P., Zhou, W., and Li, H.
\newblock Attentive experience replay.
\newblock In \emph{Proceedings of the AAAI Conference on Artificial
  Intelligence}, volume~34, pp.\  5900--5907, 2020.

\bibitem[Tang et~al.(2017)Tang, Houthooft, Foote, Stooke, Xi~Chen, Duan,
  Schulman, DeTurck, and Abbeel]{tang2017exploration}
Tang, H., Houthooft, R., Foote, D., Stooke, A., Xi~Chen, O., Duan, Y.,
  Schulman, J., DeTurck, F., and Abbeel, P.
\newblock \# exploration: A study of count-based exploration for deep
  reinforcement learning.
\newblock \emph{Advances in neural information processing systems}, 30, 2017.

\bibitem[Tassa et~al.(2018)Tassa, Doron, Muldal, Erez, Li, Casas, Budden,
  Abdolmaleki, Merel, Lefrancq, et~al.]{tassa2018deepmind}
Tassa, Y., Doron, Y., Muldal, A., Erez, T., Li, Y., Casas, D. d.~L., Budden,
  D., Abdolmaleki, A., Merel, J., Lefrancq, A., et~al.
\newblock Deepmind control suite.
\newblock \emph{arXiv preprint arXiv:1801.00690}, 2018.

\bibitem[Team et~al.(2023)Team, Bauer, Baumli, Baveja, Behbahani, Bhoopchand,
  Bradley-Schmieg, Chang, Clay, Collister, et~al.]{team2023human}
Team, A.~A., Bauer, J., Baumli, K., Baveja, S., Behbahani, F., Bhoopchand, A.,
  Bradley-Schmieg, N., Chang, M., Clay, N., Collister, A., et~al.
\newblock Human-timescale adaptation in an open-ended task space.
\newblock \emph{arXiv preprint arXiv:2301.07608}, 2023.

\bibitem[Thrun(1998)]{thrun1998lifelong}
Thrun, S.
\newblock Lifelong learning algorithms.
\newblock In \emph{Learning to learn}, pp.\  181--209. Springer, 1998.

\bibitem[Tunyasuvunakool et~al.(2020)Tunyasuvunakool, Muldal, Doron, Liu,
  Bohez, Merel, Erez, Lillicrap, Heess, and Tassa]{tunyasuvunakool2020}
Tunyasuvunakool, S., Muldal, A., Doron, Y., Liu, S., Bohez, S., Merel, J.,
  Erez, T., Lillicrap, T., Heess, N., and Tassa, Y.
\newblock dm\_control: Software and tasks for continuous control.
\newblock \emph{Software Impacts}, 6:\penalty0 100022, 2020.
\newblock ISSN 2665-9638.
\newblock \doi{https://doi.org/10.1016/j.simpa.2020.100022}.
\newblock URL
  \url{https://www.sciencedirect.com/science/article/pii/S2665963820300099}.

\bibitem[Van~Seijen et~al.(2020)Van~Seijen, Nekoei, Racah, and
  Chandar]{van2020loca}
Van~Seijen, H., Nekoei, H., Racah, E., and Chandar, S.
\newblock The loca regret: a consistent metric to evaluate model-based behavior
  in reinforcement learning.
\newblock \emph{Advances in Neural Information Processing Systems},
  33:\penalty0 6562--6572, 2020.

\bibitem[Virtanen et~al.(2020)Virtanen, Gommers, Oliphant, Haberland, Reddy,
  Cournapeau, Burovski, Peterson, Weckesser, Bright, {van der Walt}, Brett,
  Wilson, Millman, Mayorov, Nelson, Jones, Kern, Larson, Carey, Polat, Feng,
  Moore, {VanderPlas}, Laxalde, Perktold, Cimrman, Henriksen, Quintero, Harris,
  Archibald, Ribeiro, Pedregosa, {van Mulbregt}, and {SciPy 1.0
  Contributors}]{2020SciPy-NMeth}
Virtanen, P., Gommers, R., Oliphant, T.~E., Haberland, M., Reddy, T.,
  Cournapeau, D., Burovski, E., Peterson, P., Weckesser, W., Bright, J., {van
  der Walt}, S.~J., Brett, M., Wilson, J., Millman, K.~J., Mayorov, N., Nelson,
  A. R.~J., Jones, E., Kern, R., Larson, E., Carey, C.~J., Polat, {\.I}., Feng,
  Y., Moore, E.~W., {VanderPlas}, J., Laxalde, D., Perktold, J., Cimrman, R.,
  Henriksen, I., Quintero, E.~A., Harris, C.~R., Archibald, A.~M., Ribeiro,
  A.~H., Pedregosa, F., {van Mulbregt}, P., and {SciPy 1.0 Contributors}.
\newblock {{SciPy} 1.0: Fundamental Algorithms for Scientific Computing in
  Python}.
\newblock \emph{Nature Methods}, 17:\penalty0 261--272, 2020.
\newblock \doi{10.1038/s41592-019-0686-2}.

\bibitem[Wan et~al.(2022)Wan, Rahimi-Kalahroudi, Rajendran, Momennejad,
  Chandar, and Van~Seijen]{wan2022towards}
Wan, Y., Rahimi-Kalahroudi, A., Rajendran, J., Momennejad, I., Chandar, S., and
  Van~Seijen, H.~H.
\newblock Towards evaluating adaptivity of model-based reinforcement learning
  methods.
\newblock In \emph{International Conference on Machine Learning}, pp.\
  22536--22561. PMLR, 2022.

\bibitem[Waskom(2021)]{Waskom2021}
Waskom, M.~L.
\newblock seaborn: statistical data visualization.
\newblock \emph{Journal of Open Source Software}, 6\penalty0 (60):\penalty0
  3021, 2021.
\newblock \doi{10.21105/joss.03021}.
\newblock URL \url{https://doi.org/10.21105/joss.03021}.

\bibitem[{W}es {M}c{K}inney(2010)]{mckinney-proc-scipy-2010}
{W}es {M}c{K}inney.
\newblock {D}ata {S}tructures for {S}tatistical {C}omputing in {P}ython.
\newblock In {S}t\'efan van~der {W}alt and {J}arrod {M}illman (eds.),
  \emph{{P}roceedings of the 9th {P}ython in {S}cience {C}onference}, pp.\  56
  -- 61, 2010.
\newblock \doi{10.25080/Majora-92bf1922-00a}.

\bibitem[Wu et~al.(2022)Wu, Escontrela, Hafner, Goldberg, and
  Abbeel]{wu2022daydreamer}
Wu, P., Escontrela, A., Hafner, D., Goldberg, K., and Abbeel, P.
\newblock Daydreamer: World models for physical robot learning.
\newblock \emph{arXiv preprint arXiv:2206.14176}, 2022.

\bibitem[Xie et~al.(2021)Xie, Harrison, and Finn]{xie2021deep}
Xie, A., Harrison, J., and Finn, C.
\newblock Deep reinforcement learning amidst continual structured
  non-stationarity.
\newblock In \emph{International Conference on Machine Learning}, pp.\
  11393--11403. PMLR, 2021.

\bibitem[Yarats et~al.(2020)Yarats, Kostrikov, and Fergus]{yarats2020image}
Yarats, D., Kostrikov, I., and Fergus, R.
\newblock Image augmentation is all you need: Regularizing deep reinforcement
  learning from pixels.
\newblock In \emph{International Conference on Learning Representations}, 2020.

\bibitem[Yarats et~al.(2021)Yarats, Fergus, Lazaric, and
  Pinto]{yarats2021reinforcement}
Yarats, D., Fergus, R., Lazaric, A., and Pinto, L.
\newblock Reinforcement learning with prototypical representations.
\newblock In \emph{International Conference on Machine Learning}, pp.\
  11920--11931. PMLR, 2021.

\bibitem[Yu et~al.(2020)Yu, Tan, Bai, Coumans, and Ha]{yu2020learning}
Yu, W., Tan, J., Bai, Y., Coumans, E., and Ha, S.
\newblock Learning fast adaptation with meta strategy optimization.
\newblock \emph{IEEE Robotics and Automation Letters}, 5\penalty0 (2):\penalty0
  2950--2957, 2020.

\bibitem[Zenke et~al.(2017)Zenke, Poole, and Ganguli]{zenke2017continual}
Zenke, F., Poole, B., and Ganguli, S.
\newblock Continual learning through synaptic intelligence.
\newblock In \emph{International Conference on Machine Learning}, pp.\
  3987--3995. PMLR, 2017.

\bibitem[Zha et~al.(2019)Zha, Lai, Zhou, and Hu]{zha2019experience}
Zha, D., Lai, K.-H., Zhou, K., and Hu, X.
\newblock Experience replay optimization.
\newblock \emph{arXiv preprint arXiv:1906.08387}, 2019.

\bibitem[Zintgraf et~al.(2019)Zintgraf, Shiarli, Kurin, Hofmann, and
  Whiteson]{zintgraf2019fast}
Zintgraf, L., Shiarli, K., Kurin, V., Hofmann, K., and Whiteson, S.
\newblock Fast context adaptation via meta-learning.
\newblock In \emph{International Conference on Machine Learning}, pp.\
  7693--7702. PMLR, 2019.

\end{thebibliography}
\bibliographystyle{icml2023}

%%%%%%%%%%%%%%%%%%%%%%%%%%%%%%%%%%%%%%%%%%%%%%%%%%%%%%%%%%%%%%%%%%%%%%%%%%%%%%%
%%%%%%%%%%%%%%%%%%%%%%%%%%%%%%%%%%%%%%%%%%%%%%%%%%%%%%%%%%%%%%%%%%%%%%%%%%%%%%%
% APPENDIX
%%%%%%%%%%%%%%%%%%%%%%%%%%%%%%%%%%%%%%%%%%%%%%%%%%%%%%%%%%%%%%%%%%%%%%%%%%%%%%%
%%%%%%%%%%%%%%%%%%%%%%%%%%%%%%%%%%%%%%%%%%%%%%%%%%%%%%%%%%%%%%%%%%%%%%%%%%%%%%%
\newpage
\appendix
\onecolumn
\makeatletter
\let\ftype@table\ftype@figure
\makeatother

\renewcommand{\thefigure}{A\arabic{figure}}
\renewcommand{\theHfigure}{A\arabic{figure}}
\setcounter{figure}{0}   % Comment this if you don't want appendix figure numbering to start at 1

\renewcommand{\thetable}{A\arabic{table}}
\renewcommand{\theHtable}{A\arabic{table}}
\setcounter{table}{0}   % Comment this if you don't want appendix figure numbering to start at 1

\section{Societal Impact}

This work has the potential for wide-ranging applications in autonomous systems such as robotics and decision systems. Such systems offer the potential for many benefits, including with manufacturing assistance, healthcare, transportation, and scientific discovery. However, they also have the potential for negative societal impact, such as in autonomous weapons or workforce displacement. Moreover, they present risks if they are allowed to freely explore and interact with the environment, and safeguards must be put in place to ensure that such interactions do not harm humans and animals. Additionally, consideration must be made to protect against biases in an agent's learned model of the world that may not align with human values. Developing methods to test for and potentially correct such biases may thus also be an important line of research. 

\section{Extended Related Work}
\paragraph*{Model-based RL}
At least two key advantages are promised by model-based reinforcement learning relative to model-free RL: more sample efficient training due to the compression of experience into a predictive world model, and more effective policies that can leverage the world model for planning. Recent advances have propelled model-based RL to success on classic board games \cite{schrittwieser2020mastering} and on tasks with high dimensional input \cite{ha2018recurrent} including benchmarks such as Deepmind Control Suite \cite{hafner2019dream} and Atari \cite{kaiser2019model, hafner2020mastering}. These successes have been extended to robotics settings \cite{wu2022daydreamer}, goal-conditioning \cite{mendonca2021discovering}, hierarchical planning \cite{hafner2022deep}, use of a transformer architecture for improved sample-efficiency \cite{micheli2022transformers} and memory-based reasoning \cite{chen2022transdreamer}, and use of prototypes for enhanced robustness against distractions \cite{deng2022dreamerpro}. Remaining challenges include fast adaptation, and improved model performance.

\paragraph*{Adaptation}
Quick adjustment in response to changing tasks, data distributions, or environment conditions can be a very challenging problem, and is a component of continual learning \cite{thrun1998lifelong, khetarpal2022towards, julian2020never, xie2021deep}. In robotics, system identification can be used to inform adaptation to a changing environment \cite{kumar2021rma} or body \cite{bongard2006resilient, cully2015robots}. Metalearning methods \cite{finn2017model} train a model across a variety of situations such that it can adapt quickly to a new task through few-shot fine-tuning. In this approach, by seeing many different examples of how an environment might change, the agent can develop strategies for adapting and focusing on the essential aspects of an environment \cite{al2017continuous, song2020rapidly, harrison2020continuous, nagabandi2018deep, finn2019online, javed2019meta, zintgraf2019fast, yu2020learning}. Self-supervised objectives can be used to guide policy adaptation in changing environments \cite{hansen2020self, bodnar2020geometric, balloch2023neuro}. Changepoint detection \cite{adams2007bayesian, fearnhead2007line, hadoux2014sequential, banerjee2017quickest}  and context detection \cite{da2006dealing} can be used to directly signal that adaptation is required. Considerations have also been made towards the problem of catastrophic forgetting, wherein a model may lose capabilities it had previously learned \cite{kirkpatrick2017overcoming, li2017learning, zenke2017continual, rolnick2019experience, buzzega2021rethinking, ritter2018been, rusu2016progressive}. Robustness to shifting data distributions is also a problem in supervised and self-supervised learning, and benchmarks such as WILDS have been established to encourage progress \cite{koh2021wilds, sagawa2021extending}.

\paragraph*{Prioritized replay}
Experience replay consists of storing experienced episodes in a buffer and revisiting them during model updates \cite{lin1992self}. This increases efficiency by enabling data reuse, and aids training by breaking correlations that would arise in the consecutive data samples of a strictly online approach \cite{mnih2015human}. Uniform sampling from the replay buffer is a common approach \cite{hafner2019dream}. However, it is well known that prioritized sampling can yield benefits, with emphasis on prioritizing by the change in value \cite{moore1993prioritized}. In model-free contexts, there has been particular success prioritizing by the magnitude of the temporal-difference error \cite{schaul2015prioritized, horgan2018distributed, hessel2018rainbow}. Alternatively, more frequent states under the current policy can be prioritized \cite{novati2019remember, sinha2022experience, sun2020attentive}. Prioritization can also aim to minimize regret \cite{liu2021regret}, error in the value function \cite{kumar2020discor}, or where the value function is more difficult to predict  \cite{pan2020frequency}. Variations in the sampling strategy can account for the context of data relative to the rest of the replay buffer \cite{oh2020learning} or can mitigate deficiencies of outdated priorities and insufficient sample space coverage \cite{pan2022understanding}. The sampling strategy itself can even be optimized \cite{zha2019experience, burega2022learning}. In model-based contexts, states that the model can confidently make predictions about can be prioritized \cite{abbas20a}.
Particularly relevant and complementary to Curious replay is the work of \cite{oh2021model}, which investigates the use of model error for prioritization. They augment the Q-function with a parameter-sharing one-step state prediction, and uses the error of that prediction to aid prioritization. This is similar to the adversarial component of our Curious Replay, but differs in that the model itself is not used for planning. They also do not combine model error with a count-based novelty, which we find to be particularly important.   Perhaps most importantly, we demonstrate the utility of a model-loss-based prioritization for improving adaptation performance in the face of changing environments, which is not something they investigate. Moreover, while they rely on modifying the critic network to add a transition model, our approach does not require such a modification and instead uses the loss computed directly during world model updates of model-based agents. Curious Replay is explicitly targeted at improving algorithms like Dreamer that are fundamentally reliant on a world model.

\paragraph*{Intrinsically-motivated RL}
Intrinsically-motivated reinforcement learning has emerged as a strategy for guiding exploration in sparsely-rewarded environments \cite{oudeyer2007intrinsic}. For many such methods, world models play a central role in computing the intrinsic motivation signal \cite{schmidhuber1991possibility}. Such signals can include model error \cite{stadie2015incentivizing} as with ICM \cite{pathak2017curiosity} and BYOL-Explore \cite{guo2022byol}; surprise \cite{achiam2017surprise}; model learning progress as with $\gamma$-progress \cite{kim2020active}; model uncertainty as with ensemble disagreement \cite{pathak2019self, sekar2020planning}; novelty of latent state representations as with RIDE \cite{raileanu2020ride}; and information gain \cite{still2012information, houthooft2016vime}. Many signals do not require an explicit world model, such as novelty as with pseudocounts of states \cite{bellemare2016unifying, machado2020count} and changes \cite{parisi2021interesting}, random network distillation \cite{burda2018exploration}, state-hashing  \cite{tang2017exploration}, entropy with prototypical representations \cite{yarats2021reinforcement}; diversity \cite{eysenbach2018diversity}; and empowerment \cite{klyubin2005all, mohamed2015variational, gregor2016variational}.

\section{Validation of expected object interaction behavior with mice}
We validated the expected animal behavior with C57Bl/6 mice (Black 6, Jackson Laboratory, 664). Mice were allowed to freely investigate a 12" x 12" white acrylic box. In the unchanging environment condition, a small ($\sim$0.5" diameter) pink rubber ball was present in the box from the outset, before the mouse was placed in the box. In the changing environment condition, the ball was placed in the box (using metal tongs) after 10 minutes of exploration of the empty box. In both conditions, the mouse was allowed to explore the box while the ball was present for ten minutes. The behavior was recorded using a webcam, and the time of each interaction between the mouse and ball was manually scored. All procedures were performed in accordance with protocols approved by the Stanford University Institutional Animal Care and Use Committee (IACUC) and guidelines of the National Institutes of Health. 

\begin{figure}
\centerline{\includegraphics[width=.5\textwidth]{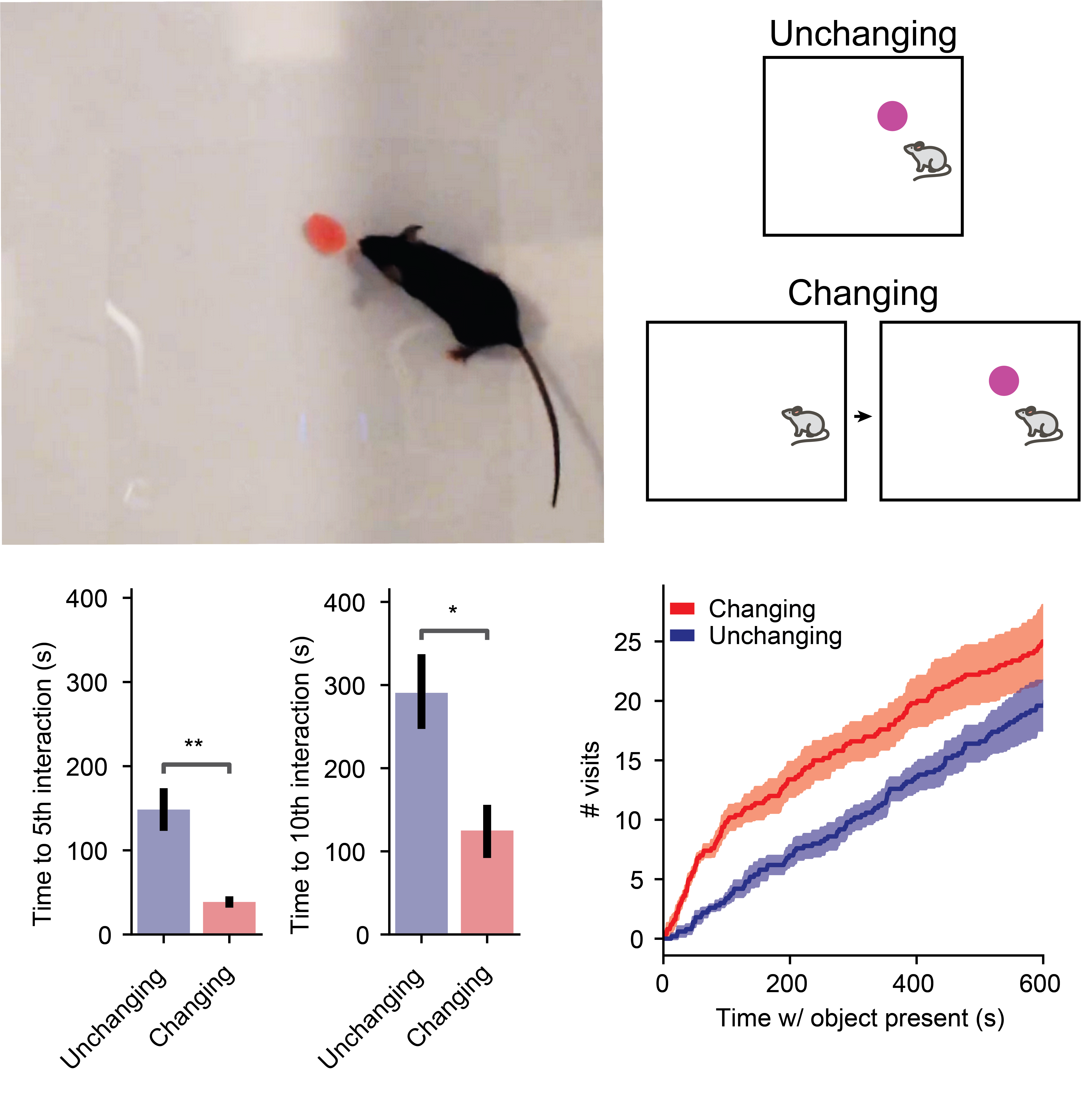}}
\caption{Novel-object interaction of wildtype mice in changing and unchanging environments. The 5th and 10th interactions both occur after significantly less time in the changing environment (n=5 each condition, independent t-test, * $p < 0.05$, ** $p < 0.001$).}
\label{figS1}
\end{figure}

\begin{figure}
\centerline{\includegraphics[width=.5\textwidth]{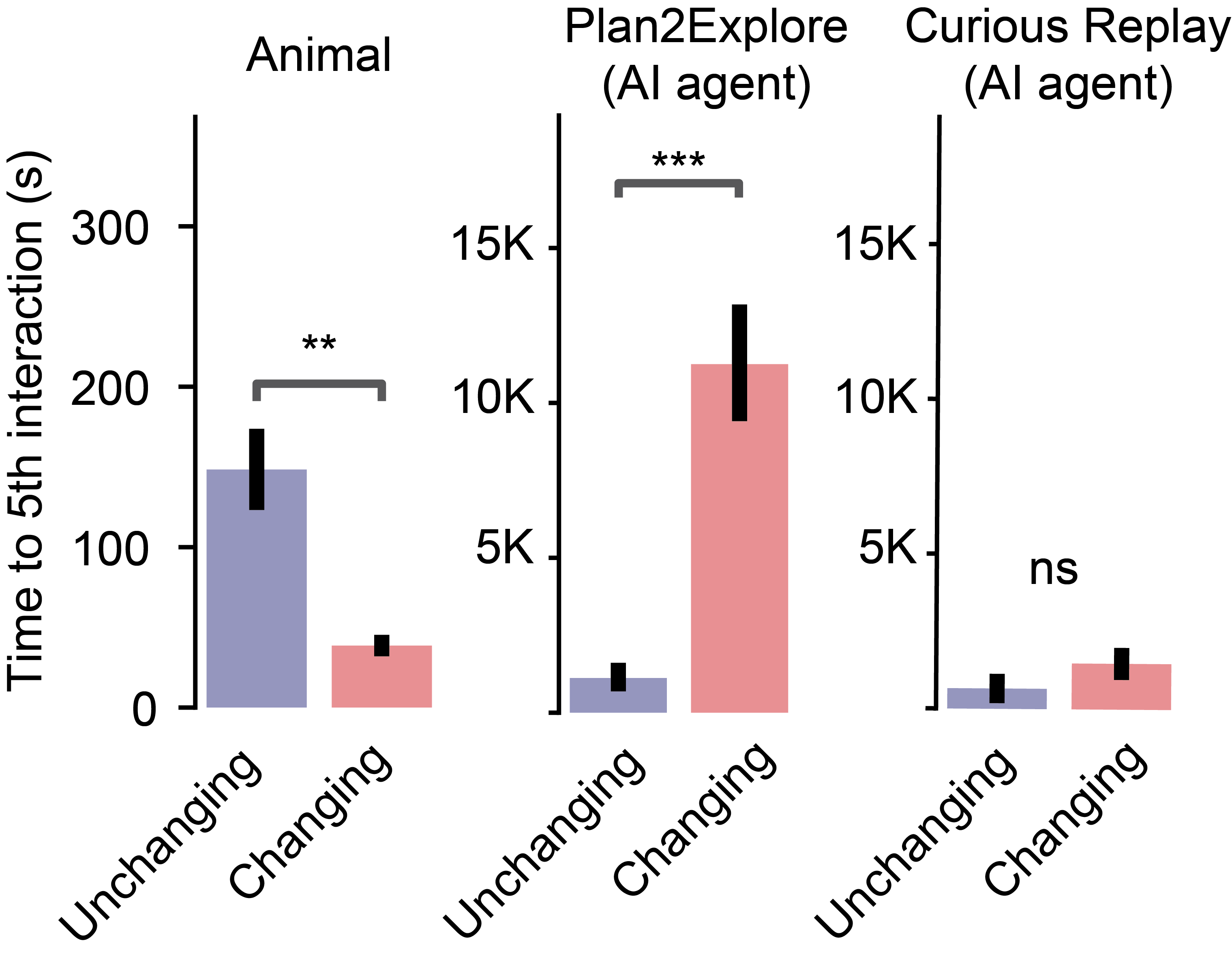}}
\caption{
Comparison of animal (mouse) behavior and AI agent behavior. 
Baseline Dreamer AI agent has the opposite behavior as animals, with a longer time to interaction in the changing environment. Curious Replay reduces this gap in behavior, bring behavior in the changing environment on par with in the unchanging environment. There still remains a gap, however, that future work can investigate. AI agent steps have been converted into seconds using the 0.03 seconds/step timescale of the simulated environment. (Animal: n=5 each condition,  Baseline: n=7 each condition, Curious Replay: n=7 each condition,  independent t-test, * $p < 0.05$, ** $p < 0.001$).}
\label{figS0}
\end{figure}

\section{Technical details}

\subsection{Implementation details} We use the STArr python package \href{https://github.com/justinmaojones/starr} as a fast SumTree implementation for DreamerV2 and DreamerPro. For Curious Replay, a running minimum (across the entire run) was subtracted from the loss before a priority value was computed. We did not use the running minimum when prioritizing with temporal difference. Dreamer is implemented in Tensorflow2. We customize the MuJoCo-based dm\_control library for object interaction assay implementation, with a control timestep of 0.03 s and a physics simulation timestep of 0.005 s. For the Background-Swap Control Suite, a ground-plane alpha of 0.1 is used. Code will be made publicly available. 

In DreamerV3, sequences of length 64 are stored in the replay buffer. In the Curious Replay implementation for DreamerV3, the probability of training on a sequence is based on the priority calculated for the last step of the sequence. The Reverb replay buffer \cite{cassirer2021reverb} is used to store the sequences and priorities and to select the samples. After each training step, the training count and priority for each step in the sequence is updated. Unlike in the DreamerV2 implementation, the loss used to calculate the priority is not adjusted by the running minimum.

Episode length is 1K steps for Control Suite and 100K steps for object interaction to allow for substantial uninterrupted exploration. In Crafter, episode length depends on the agent's survival. An action repeat of 2 \cite{hafner2019dream} is applied across object interaction and Control Suite environments, with no action repeat for Crafter.

\subsection{Detailed hyperparameters}
For object interaction and Control Suite, batches are $B=10$ sequences of fixed length $L=50$. 
 For continuous control tasks, we used a nondiscrete latent space, which we found performed better. For Crafter, we used a slightly more updated version of DreamerV2 that includes layer normalization, $B=16$ and a discrete latent space. Model training and evaluation used Google Cloud T4 GPU instances. Each agent’s online return is logged every episode for Control Suite, and every 20 steps for object interaction.

\textbf{Object interaction assay and Control Suite}
    The Recurrent State Space Machine (RSSM) is nondiscrete with a  200 unit hidden state, a GRU with 200 units, and a stochastic state with 32 units. The convolutional image encoder has 4 layers with depth 48, kernel sizes of 4, and an output of size 512. The convolutional image decoder has 4 layers of depth 48, kernels of size [5, 5, 6, 6], and an output of size 64 x 64 x 3. The reward prediction head, actor, and critic are each an MLP with 4 layers of 400 units. The world model is trained using just the image gradient. The actor is trained using straight-through dynamics backpropagation. World model optimization uses Adam with learning rate of 3e-4 and epsilon 1e-5, weight decay 1e-6, and clipping of 100,  and actor-critic optimization has a learning rate of 8e-5. The models are trained every 5 environment steps. 
    Plan2Explore uses a disagreement ensemble of 10 action-conditioned one-step MLP models with 4 layers each with 400 units. The disagreement target is the stochastic state. Ensemble optimization uses Adam with learning rate 3e-4 and epsilon 1e-5.

\textbf{Crafter}
The settings for Crafter are the same, except that the RSSM is discrete with 1024 unit hidden layer, 1024 unit GRU, and 32 x 32 dimensional discrete stochastic state. Layer normalization is also applied. Optimization of the world model and actor critic use Adam with a learning rate of 1e-4, epsilon 1e-5, weight decay 1e-6, and clipping of 100. 

\section{Additional discussion of Curious Replay}
\begin{figure}[t]
\centerline{\includegraphics[width=.5\textwidth]{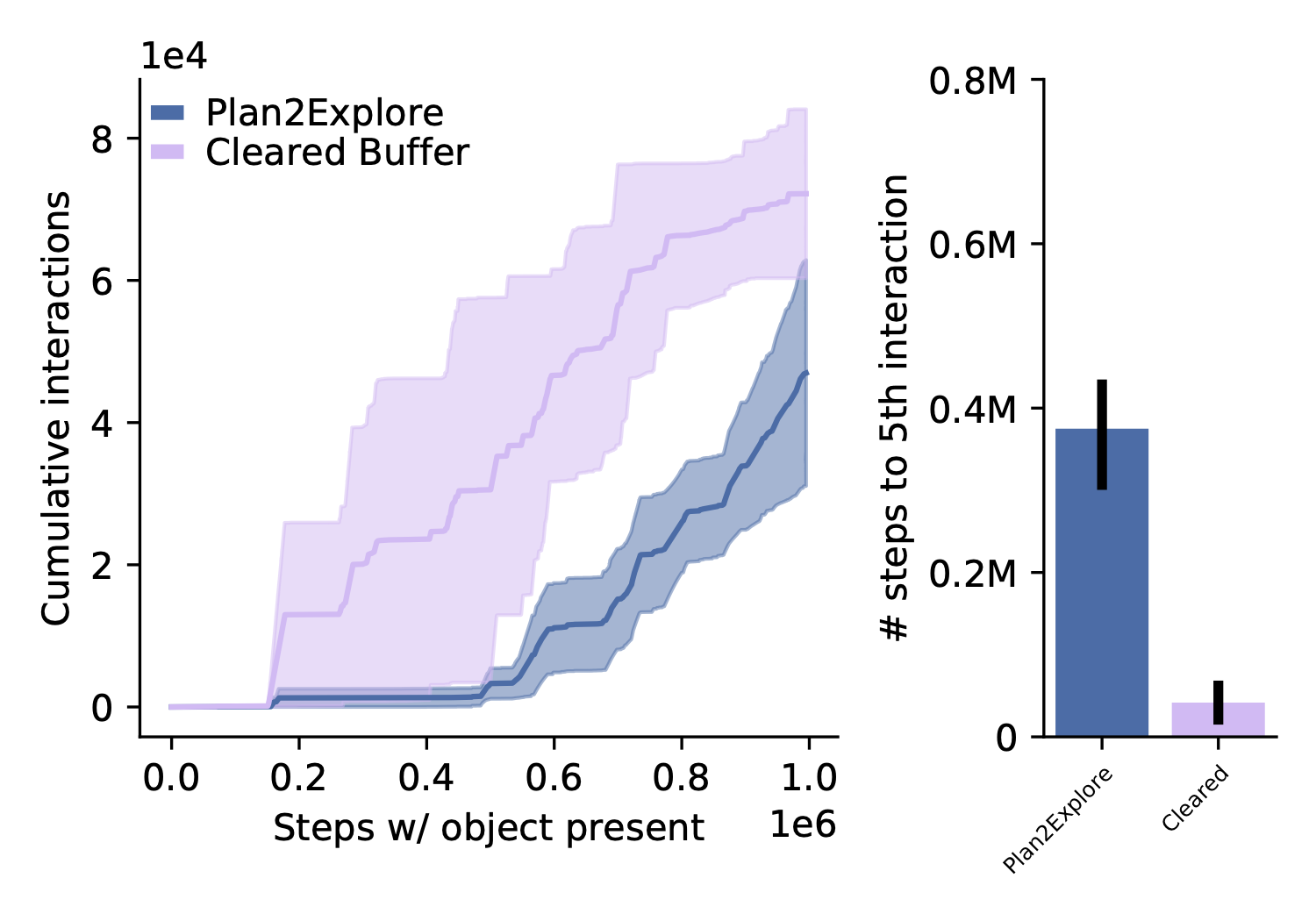}}
\caption{Clearing the replay buffer at the time the environment changes (step 5e5) substantially improves the time to object interaction. (Baseline n=7, Cleared buffer n=2).}
\label{fig_clearbuffer}
\end{figure}

We tested the hypothesis that Dreamer's uniform replay buffer sampling does not promote training the world model on new data from the changed environment. In the object interaction assay, we cleared the replay buffer at $T_0$, when the environment changed. Clearing the buffer, which forces the model to train only on new data, substantially improves the time to object interaction (Figure \ref{fig_clearbuffer}), suggesting that a key issue is how data is sampled from the replay buffer. However, clearing the replay buffer is a problematic approach. In particular, it is important to not be trigger-happy in clearing the buffer, and thus high accuracy for detecting changepoints is required. Moreover, even with accurate changepoint detection, prematurely clearing valuable data may lead to catastrophic forgetting or other issues. We thus sought a solution with the performance benefits of clearing the buffer but with more generality and less brittleness. Count-based Replay is one such solution, inspired by count-based exploration \cite{bellemare2016unifying}.

\section{Method comparisons}
For comparison to other methods, we used the official implementations corresponding to the publications. For DreamerV2, we forked \href{https://github.com/danijar/dreamerv2/tree/3a711b42461b9942396f84ad3a63ec00f25faedb}{commit 3a711b4}. For DreamerPro, we forked \href{https://github.com/fdeng18/dreamer-pro/tree/809b41757e1b08322011d5bcfc6e9f577f82c138}{commit 809b417}. For IRIS, we forked \href{https://github.com/eloialonso/iris/tree/e6aaa67a98474a9b9888b733b3eb40fb5ca9401a}{commit e6aaa67}. For DreamerV3, we forked \href{https://github.com/AutonomousAgentsLab/curiousreplay-dv3/commit/84ecf191d967f787f5cc36298e69974854b0df9c}{commit 84ecf191}.

\section{Helpful python libraries}
We thank the creators of matplotlib \cite{Hunter:2007}, pandas \cite{mckinney-proc-scipy-2010}, seaborn \cite{Waskom2021}, scipy \cite{2020SciPy-NMeth}, numpy \cite{harris2020array}, pillow \cite{clark2015pillow}, and tensorflow \cite{tensorflow2015-whitepaper}. 

\pagebreak

\begin{figure}[t]
\centerline{\includegraphics[width=.5\textwidth]{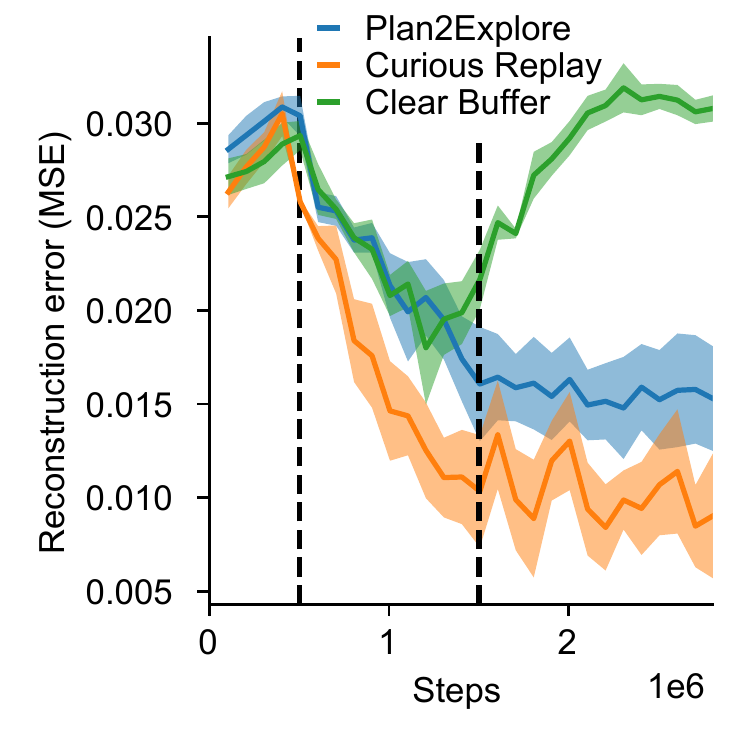}}
\caption{Testing catastrophic forgetting of the world model. Here, the object is introduced at step 0.5e6 and removed at step 1.5e6. Model performance is assessed using a test suite of trajectories that include the object. There are clear signs of catastrophic forgetting when clearing the replay buffer (at step 1.5e6) but not with Curious Replay (n=6, mean $\pm$ s.e.m.).}
\label{fig_clearbuffer_error}
\end{figure}

\begin{figure}[t]
\centerline{\includegraphics[width=.5\textwidth]{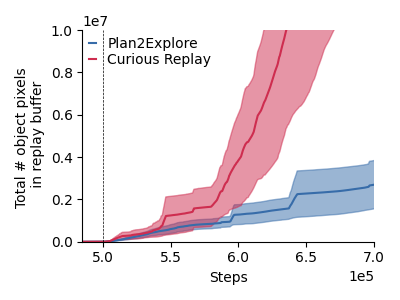}}
\caption{The agents are quickly exposed to the object once it appears in the environment. By quantifying the number of object pixels that are in the replay buffer, we see that at step 5e5 (the step when the object is introduced into the environment) the agent begins to experience observing the object (n=7, 7 for uniform and MT agents, mean +/- s.e.m.). For the first 5e4 steps, Curious Replay and Plan2Explore experience the object similarly, but diverge as Curious Replay begins to interact with the object sooner.}
\label{fig_npix_magenta}
\end{figure}

\begin{figure}[t]
\centerline{\includegraphics[width=.5\textwidth]{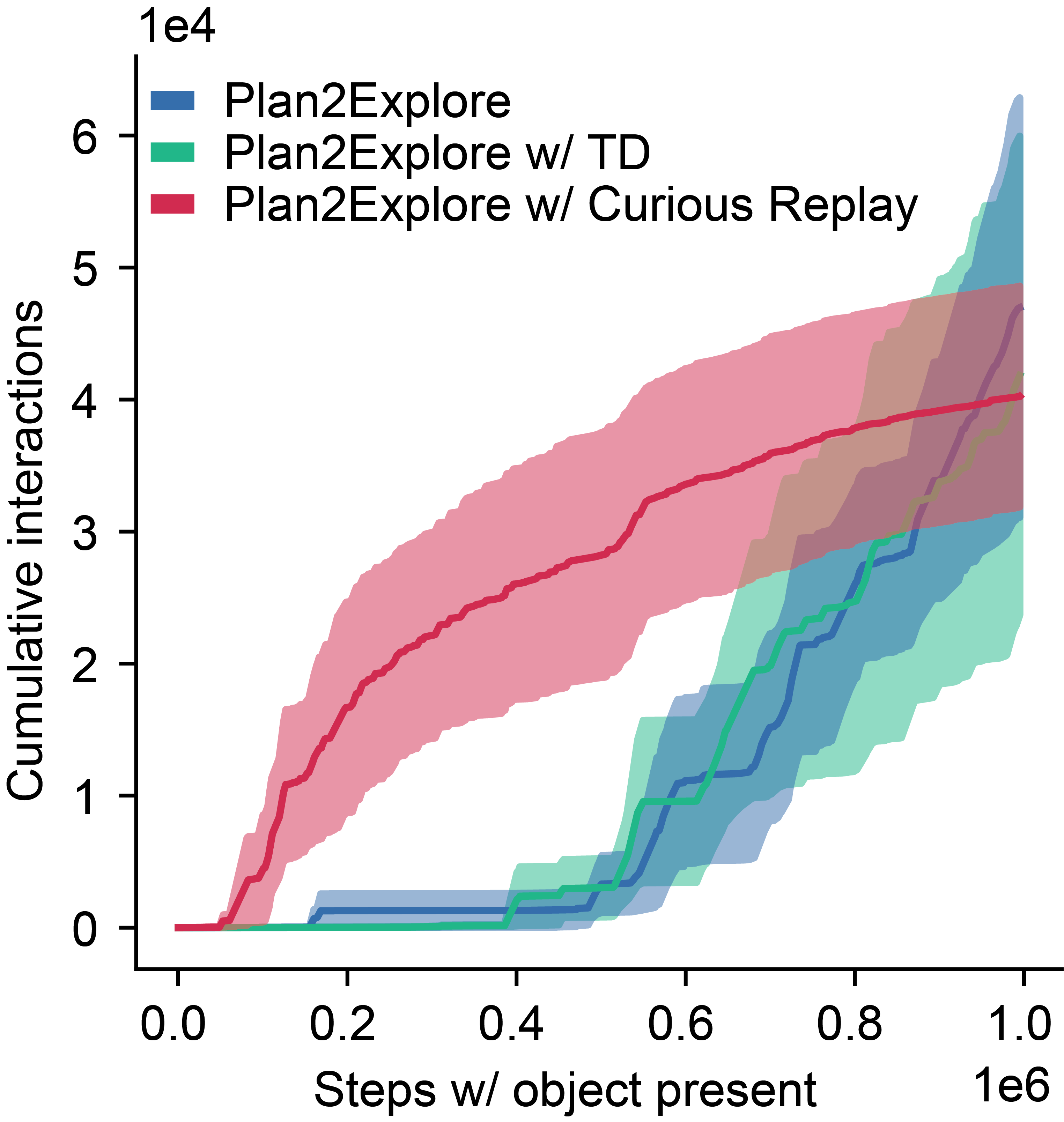}}
\caption{Cumulative object interactions over time. (n=7 each condition, mean +/- s.e.m.).}
\label{fig_cumulative}
\end{figure}

\begin{figure}[t]
\centerline{\includegraphics[width=.5\textwidth]{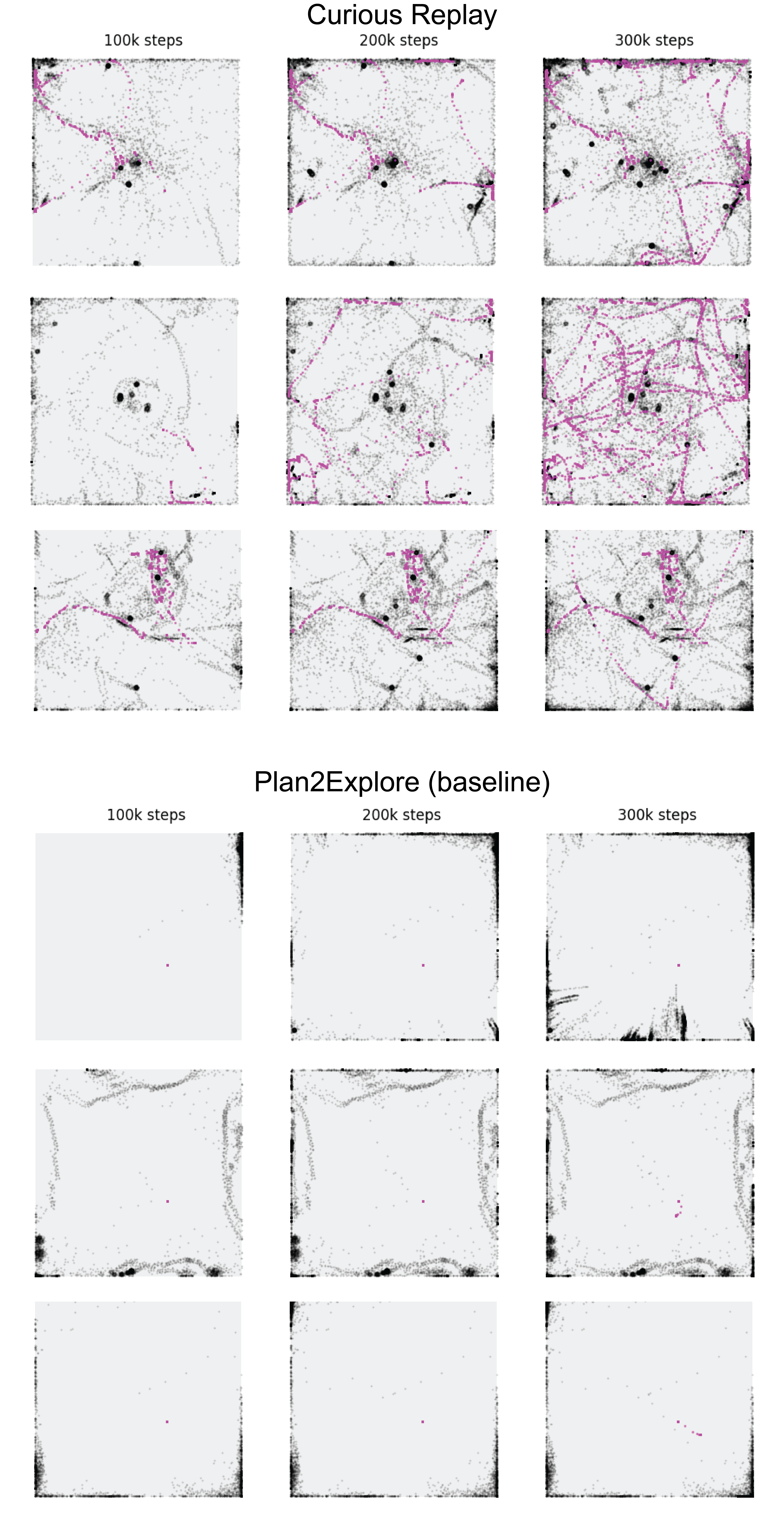}}
\caption{Example baseline Dreamer and Curious Replay agent trajectories at 100K, 200K, and 300K steps after the object has been introduced. Black dot represents the location of the agent, and magenta represents the location of the object.}
\label{fig_example_trajectories}
\end{figure}

\begin{figure}[t]
\centerline{\includegraphics[width=.4\textwidth]{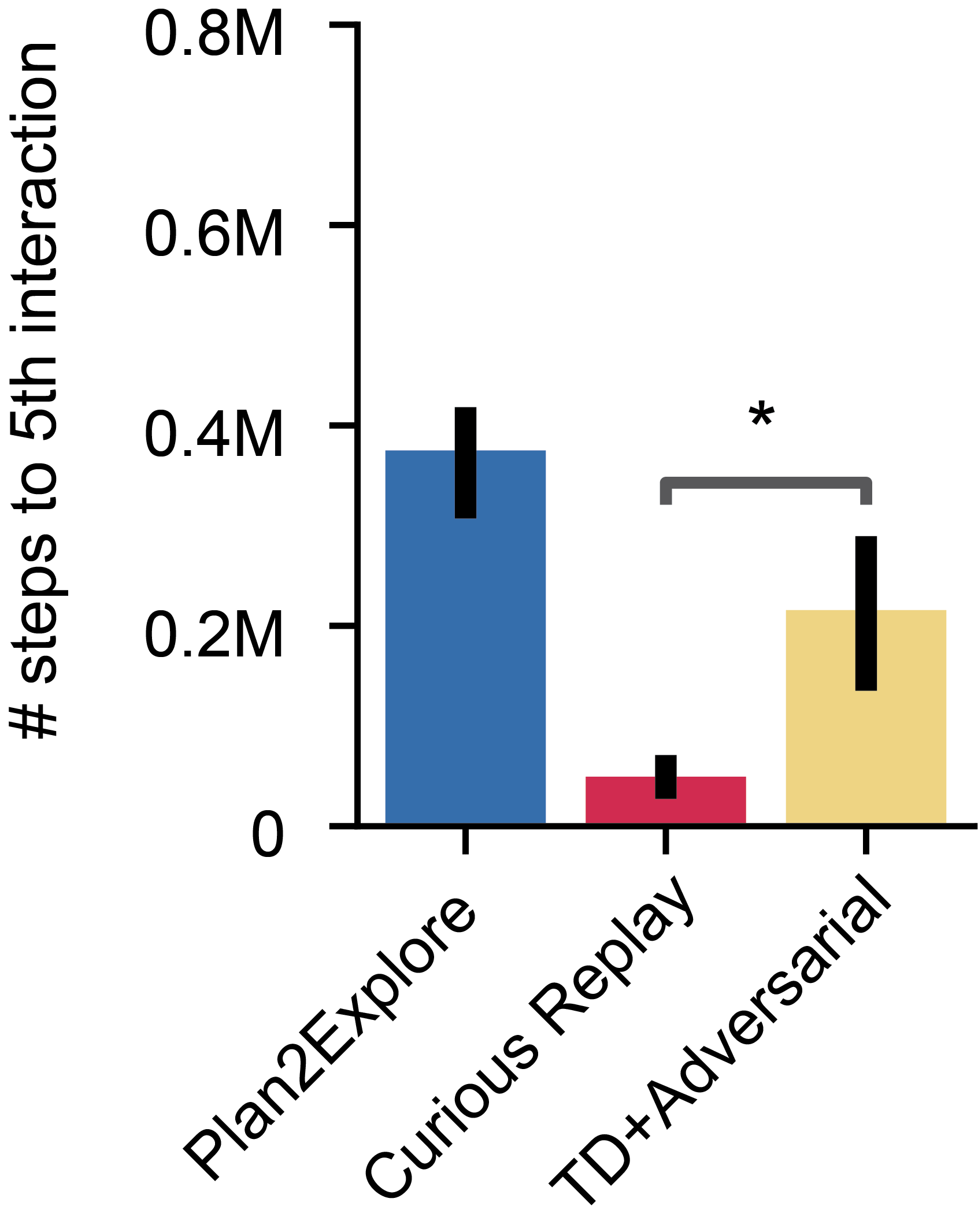}}
\caption{Comparison of Curious Replay with combined TD + Adversarial prioritization. To combine TD and Adversarial priorities while accounting for potentially very different scaling of the priority values, two separate priority arrays were used, and experiences were sampled according to each prioritization, with a fraction $f \in [0, 1]$ experiences according to the TD prioritization, and $(1-f)$ according to the Adversarial prioritization. Results shown here investigate a range of $f$ values from 0.1 to 0.9, none of which yielded a result on par with Curious Replay (n=7 Curious Replay, n=6 total TD+Adversarial, mean +/- s.e.m., independent t-test).}
\label{md_comparison}
\end{figure}

\begin{figure}[t]
\vskip 0.1in
\begin{center}
\centerline{\includegraphics[width=0.5\textwidth]{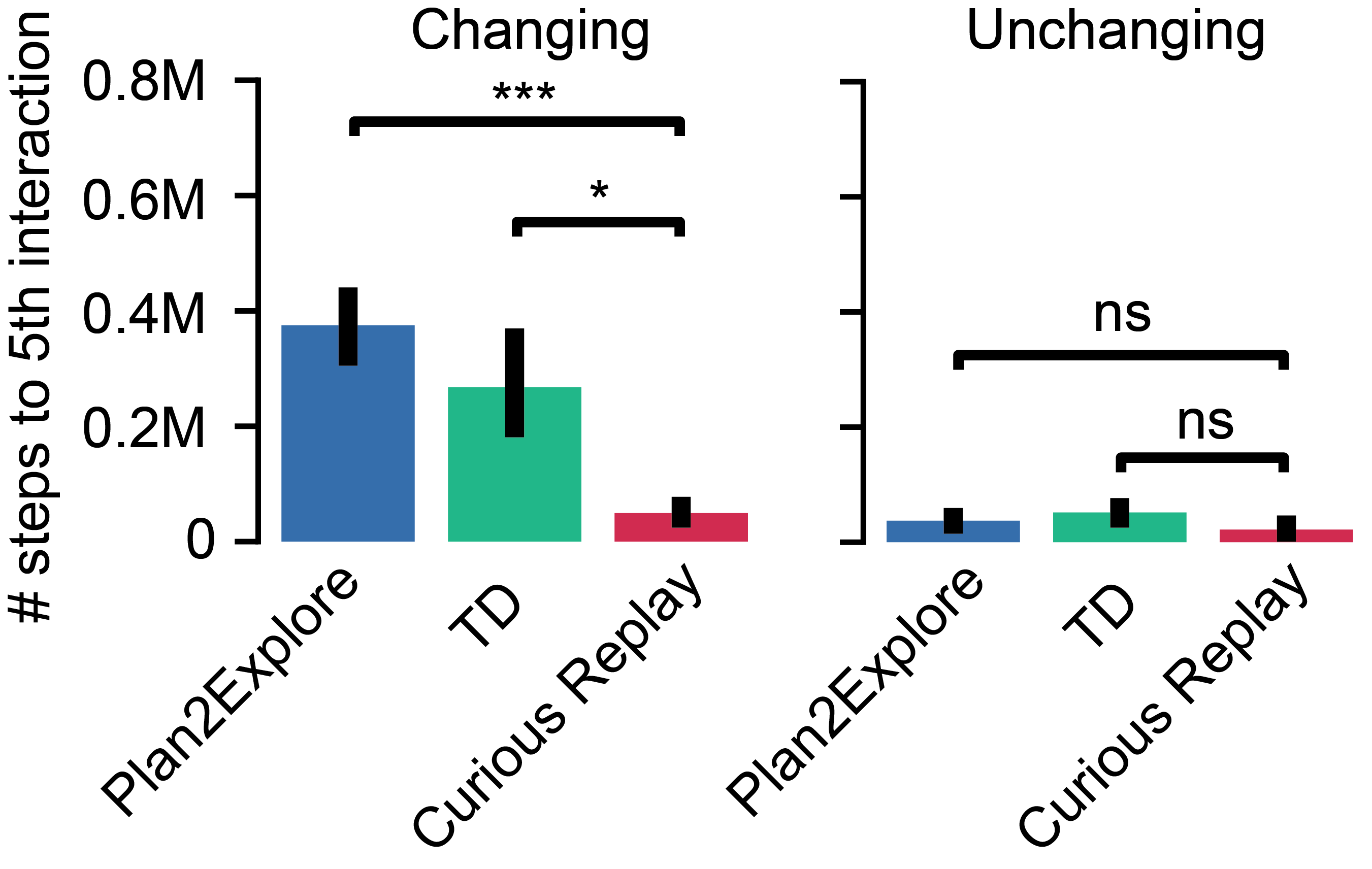}}
\vskip -0.1in
\caption{Object interaction assay, comparing Plan2Explore, Plan2Explore with temporal-difference prioritization (TD), and Plan2Explore with Curious Replay (ours). Curious Replay is significantly quicker in the changing environment. In the unchanging environment, Curious Replay remains as fast as Plan2Explore (n=7 each condition, independent t-test with fdr-bh correction). 
}
\label{s1s2}
\end{center}
\vskip -0.4in
\end{figure}

\begin{figure}[t]
\centerline{\includegraphics[width=.9\textwidth]{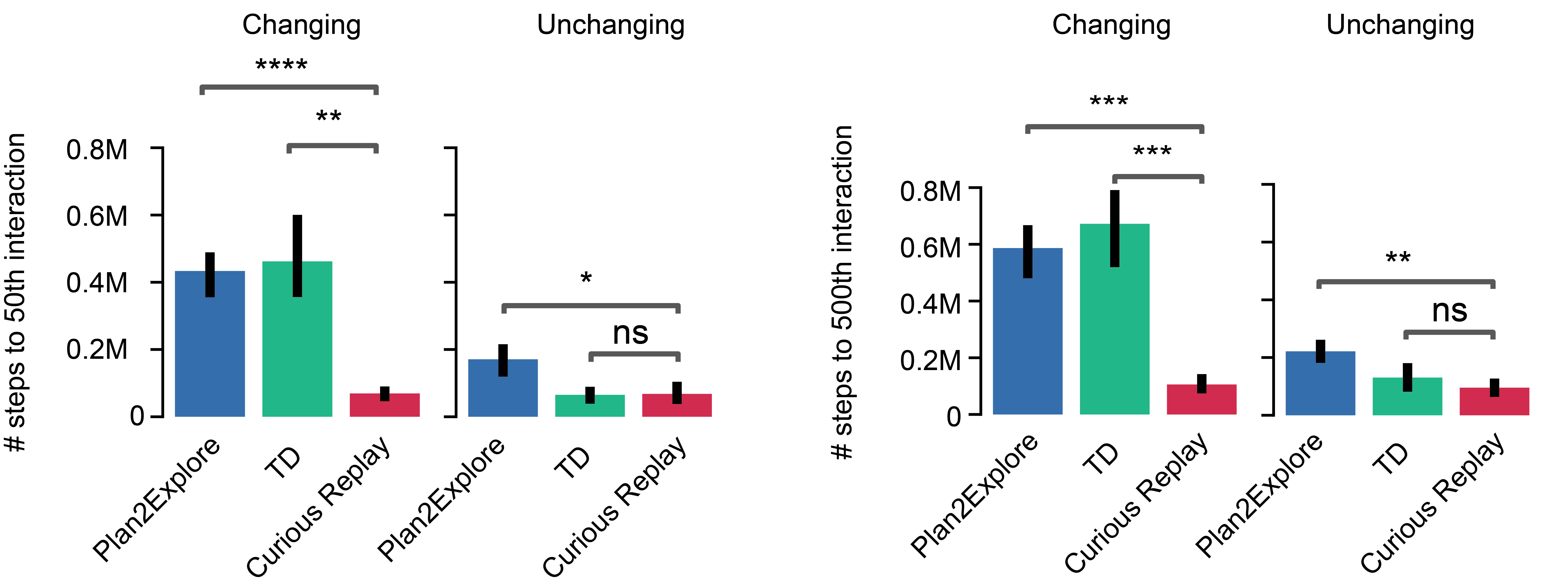}}
\caption{Robustness of Curious Replay performance to measurement by different metrics of object interactions (time to 50th or 500th interaction). (n=7 each condition, mean +/- s.e.m., independent t-test with fdr-bh correction).}
\label{fig_Nth}
\end{figure}

\begin{figure}[t]
\centerline{\includegraphics[width=.8\textwidth]{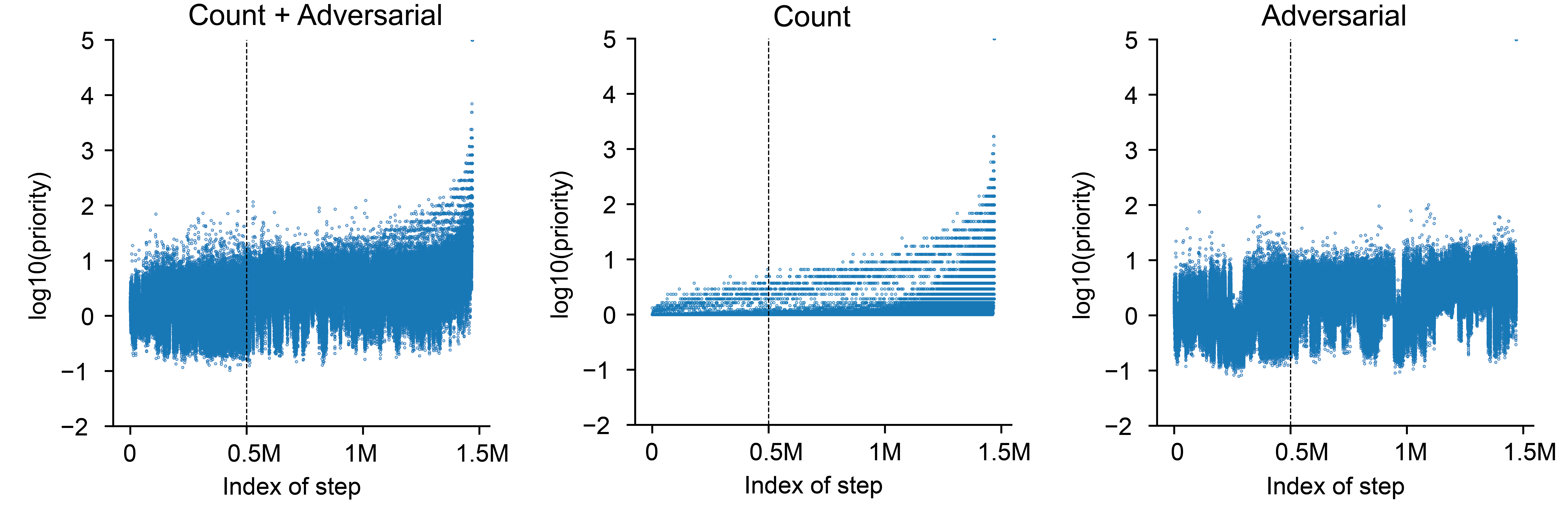}}
\caption{Example prioritizations across replay buffer containing 1.5e6 steps from playground environment (object is introduced at step 0.5e6).}
\label{fig_priorities}
\end{figure}

\begin{figure}[t]
\centerline{\includegraphics[width=.4\textwidth]{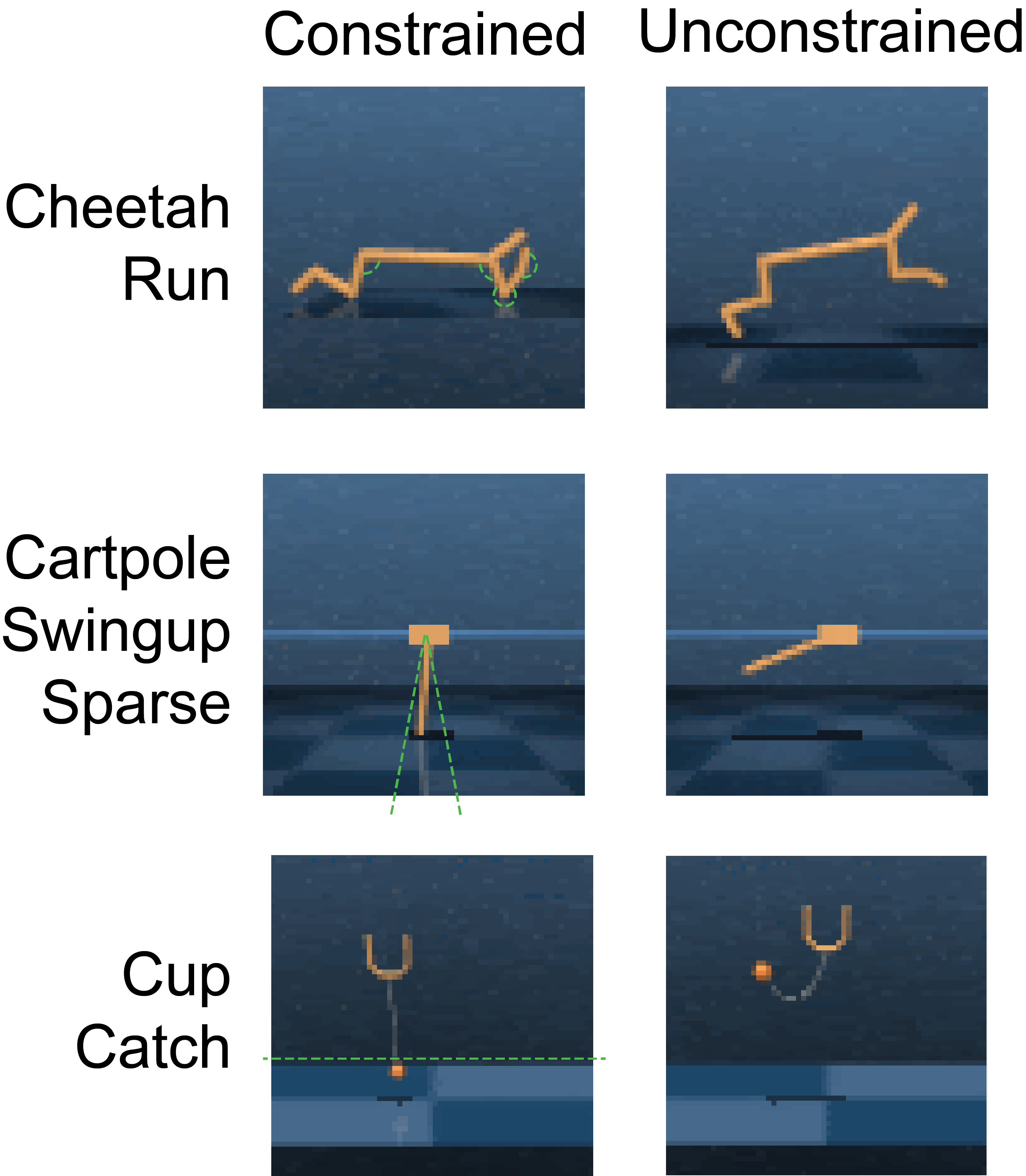}}
\caption{Constrained Control Suite tasks. Green dashed lines represent the constraints. For cup, the z-range of the ball was limited to [-.151 -.091]. For cartpole, the hinge of the pole was limited to angles [175 185]. For cheetah, the limits of the joints were bthigh [50 60], fthigh [-57 -52], fshin [-70 -60], ffoot [-28 -26]. In all cases, at $T_0=500K$, all constraints were released. }
\label{fig_cdmc}
\vskip 0.5in
\end{figure}

\begin{table}
\caption{Hyperparameter optimization on object interaction assay (n=2 seeds per setting, mean $\pm$ s.e.m.). The meaningful hyperparameters are $\alpha$
 and $\beta$
 (see Equation \ref{p_eqn}, which control the degree of adversarial prioritization, and the fall-off rate of count-based prioritization, respectively). The hyperparameter $c$
 can further trade off the relative importance of the two prioritizations; we set it to scale the count-based signal to be slightly stronger than the adversarial signal (as informed by the scale of the average model loss, here we set $c=1e4$). We set $\epsilon$
 to 0.01 and never modified, and $maxval$
 was set such that it was greater than any other priority. Thus $\alpha$
, $\beta$, and potentially $c$ are the only hyperparameters that require attention. We optimized the hyperparameters in the object-interaction assay, as we were developing the method. Due to computational cost, we did not do a full grid search. We anchored our search around 
$\alpha$=0.7 and 
$\beta$=0.7, based on the hyperparameters in the original Prioritized Experience Replay paper. Our search varied $\alpha$ or $\beta$
, with the other value set to 0.7. We found that higher values of $\alpha$
 and $\beta$ were better, with ($\alpha$
=0.7, $\beta$=0.7) the best among settings that we tried. Importantly, all tested hyperpameter values yielded better object interaction than the Plan2Explore baseline. We thus used ($\alpha$
=0.7, $\beta$=0.7) for the rest of the experiments.}
\centering
\vskip 0.1in
\begin{tabular}{|l|c|c|c|c|c|c|}

\hline
\textbf{Value of $\boldsymbol\beta$ (with $\boldsymbol\alpha$=0.7)}	& Baseline	& 0.1	& 0.3	& 0.5	& 0.7	& 0.9 \\
\hline
$\#$ Steps to 5th interaction (x$10^5$)	& 3.7 $\pm$ 0.6	& 1.2 $\pm$ 0.9	& 1.1 $\pm$ 0.3	& 1.4 $\pm$ 0.4	& \textbf{0.5 $\pm$ 0.1}	& 0.6 $\pm$ 0.2 \\
\hline
\hline
\textbf{Value of $\boldsymbol\alpha$ (with $\boldsymbol\beta$=0.7)} & Baseline	& 0.1	& 0.3	& 0.5	& 0.7	& 0.9 \\
\hline
$\#$ Steps to 5th interaction (x$10^5$) & 	3.7 $\pm$ 0.6	& 2.5 $\pm$ 1.2 & 	1.5 $\pm$ 0.4	& 2.0 $\pm$ 0.6& 	\textbf{0.5 $\pm$ 0.1}	& 0.9 $\pm$ 0.3\\

\hline
\end{tabular}
\label{hyperparam_sensitivity}
\end{table}

\begin{table}
\caption{Hyperparameter optimization on Crafter ($10\ge n \ge 5$ seeds per setting, mean $\pm$ s.d.). To assess sensitivity to Curious Replay hyperameters, we tested a variety of hyperparameters for DreamerV2 with CR on Crafter, by varying either $\alpha$ or $\beta$
, with the other value set to 0.7. We further tested the impact of using 
$c$=1e3 instead of $c$=1e4 (the value we had used for all other experiments). We found that all tested hyperparameter settings yielded better Crafter performance than the DreamerV2 and DreamerV2 w/ TD baselines. Notably, some of the hyperparameter settings actually yielded a further improved score. Thus the improvement by CR relative to the baseline lacks hyperparameter sensitivity, and promisingly, there appears the potential for hyperparameter optimization on specific tasks to yield even better performance. In the main text, we report the optimized DreamerV2 w/ CR score ($\alpha=0.5$, $\beta=0.7$) even though all other results in the main text use a fixed ($\alpha=0.7$, $\beta=0.7$).}
\centering
\vskip 0.1in
\begin{tabular}{|l|c|}
\hline
\textbf{Method} & \textbf{Crafter Score}  \\
\hline

DreamerV2 &	11.7 $\pm$ 0.5 \\
DreamerV2 w/ TD	& 10.8 $\pm$ 0.6 \\
DreamerV2 w/ CR: $\alpha$=0.7, $\beta$=0.9, c=1e4	&11.9 $\pm$ 0.8 \\
DreamerV2 w/ CR: $\alpha$=0.7, $\beta$=0.8, c=1e4	&13.2 $\pm$ 1.9 \\
DreamerV2 w/ CR: $\alpha$=0.7, $\beta$=0.7, c=1e4	&12.0 $\pm$ 1.2 \\
DreamerV2 w/ CR: $\alpha$=0.7, $\beta$=0.6, c=1e4	&13.2 $\pm$ 1.5 \\
DreamerV2 w/ CR: $\alpha$=0.7, $\beta$=0.5, c=1e4	&12.7 $\pm$ 1.1 \\
DreamerV2 w/ CR: $\alpha$=0.9, $\beta$=0.7, c=1e4	&13.0 $\pm$ 1.7 \\
DreamerV2 w/ CR: $\alpha$=0.8, $\beta$=0.7, c=1e4	&13.2 $\pm$ 1.4 \\
DreamerV2 w/ CR: $\alpha$=0.6, $\beta$=0.7, c=1e4	&12.5 $\pm$ 1.2 \\
DreamerV2 w/ CR: $\alpha$=0.5, $\beta$=0.7, c=1e4	& \textbf{13.3 $\pm$ 1.3} \\
DreamerV2 w/ CR: $\alpha$=0.7, $\beta$=0.7, c=1e3	&12.6 $\pm$ 2.0 \\
\hline
\end{tabular}
\label{hyperparam_sensitivity_crafter}
\end{table}

\begin{table}
\caption{Increased train ratio benefits Curious Replay (n=10, mean $\pm$ s.d.). }
\centering
\vskip 0.1in
\begin{tabular}{|l|c|}
\hline
\textbf{Method} & \textbf{Crafter Score}  \\
\hline
DreamerV2 &	11.7 $\pm$ 0.5 \\
DreamerV2, 8x train frequency	& 11.0 $\pm$ 1.5  \\
DreamerV2 w/ CR ($\alpha$=0.7, $\beta$=0.7)	&12.0 $\pm$ 1.2 \\
DreamerV2 w/ CR ($\alpha$=0.7, $\beta$=0.7),	8x train frequency & \textbf{15.7 $\pm$ 2.4} \\
\hline
\end{tabular}
\label{train_ratio_table}
\end{table}

\begin{figure*}[t]
\centerline{\includegraphics[width=\textwidth]{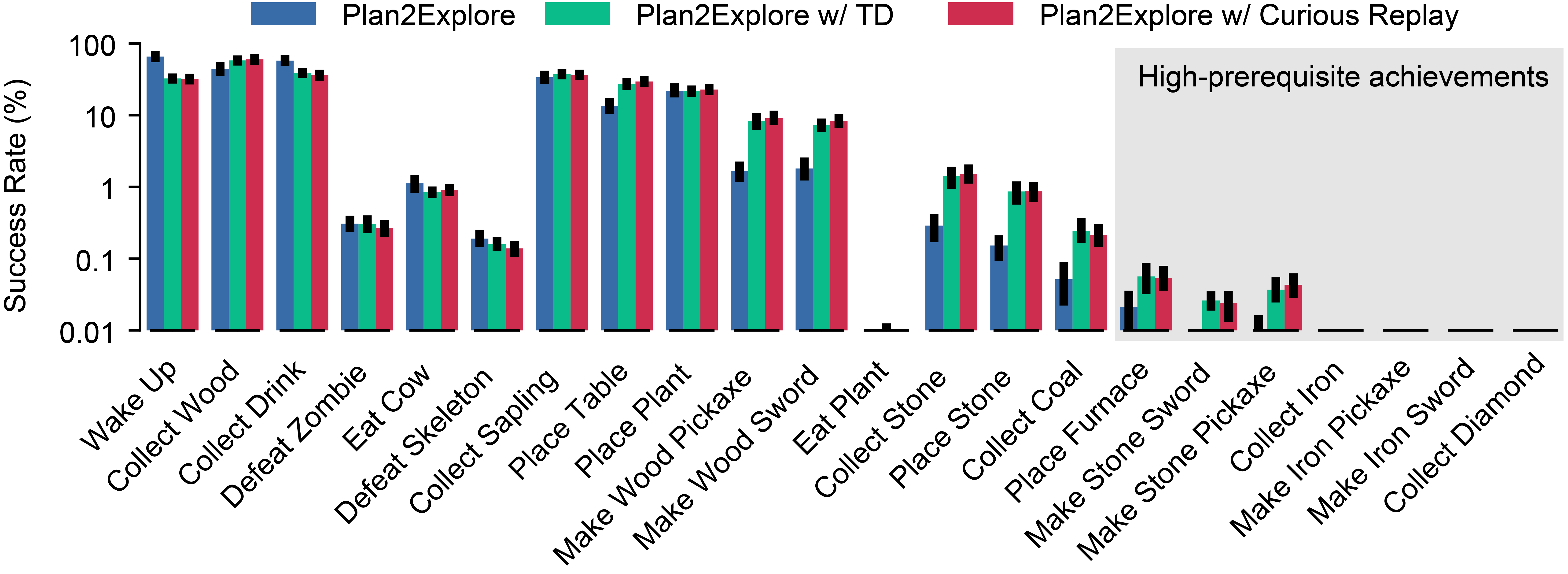}}
\caption{Agent ability spectrum showing success rates for unsupervised Crafter, ordered from left to right by number of prerequisites for an achievement (n=8 each condition,  mean $\pm$ s.e.m.). Highlighted high-prerequisite achievements need at least four preceding achievements.}
\label{fig_spectrum_unsupervised}
\end{figure*}

\begin{figure}[h]
\vspace{-15pt}
\vskip 0.2in
\begin{center}
\centerline{\includegraphics[width=0.3\textwidth]{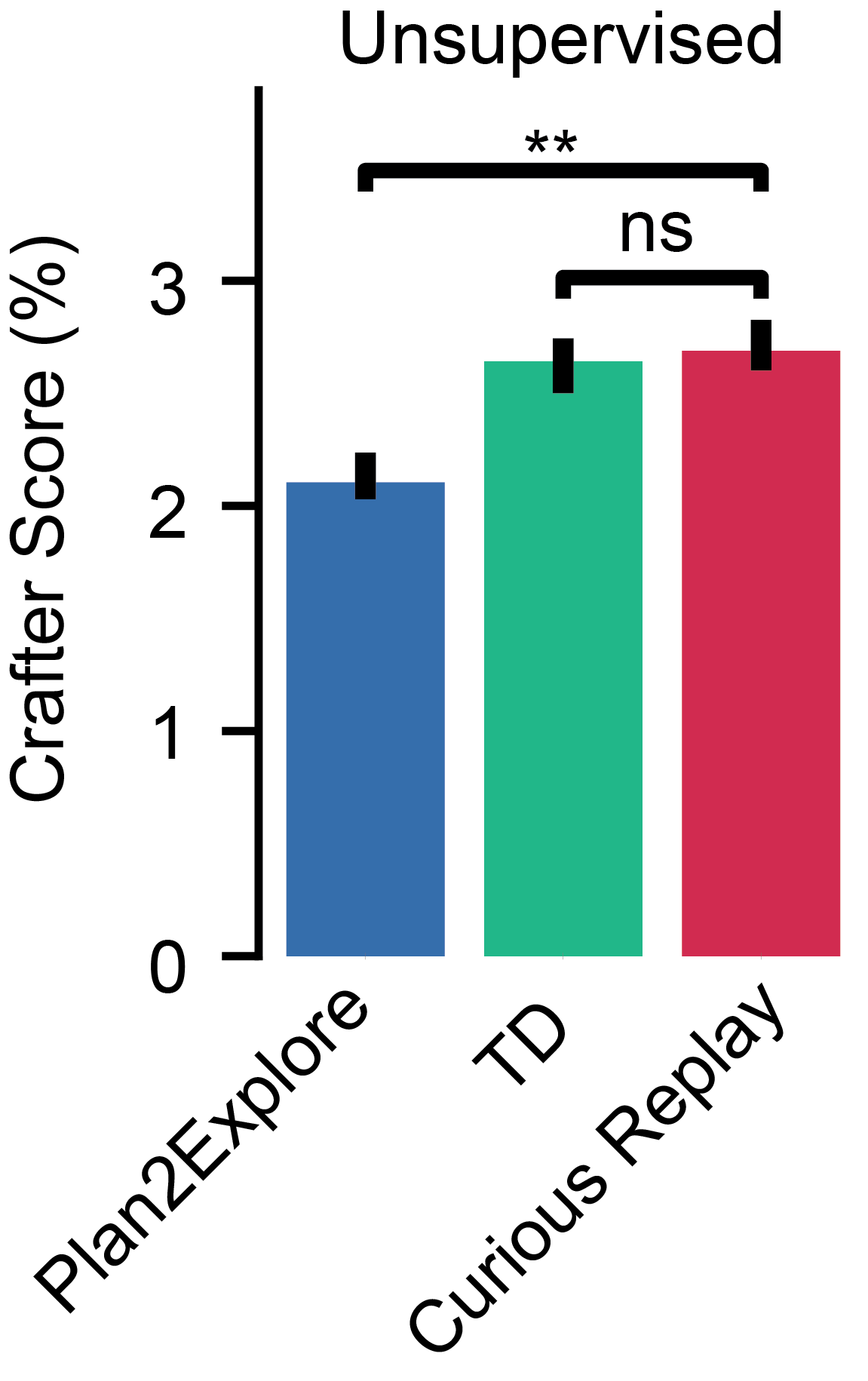}}
\vspace{-15pt}
\caption{Score on the unsupervised (no extrinsic reward) version of Crafter (n=8 each condition, mean +/- s.e.m., independent t-test with fdr-bh correction).}
\label{crafter_score_unsupervised}
\end{center}
\vskip -0.4in
\end{figure}

\begin{figure*}[ht]
\vskip 0.2in
\begin{center}
\centerline{\includegraphics[width=\textwidth]{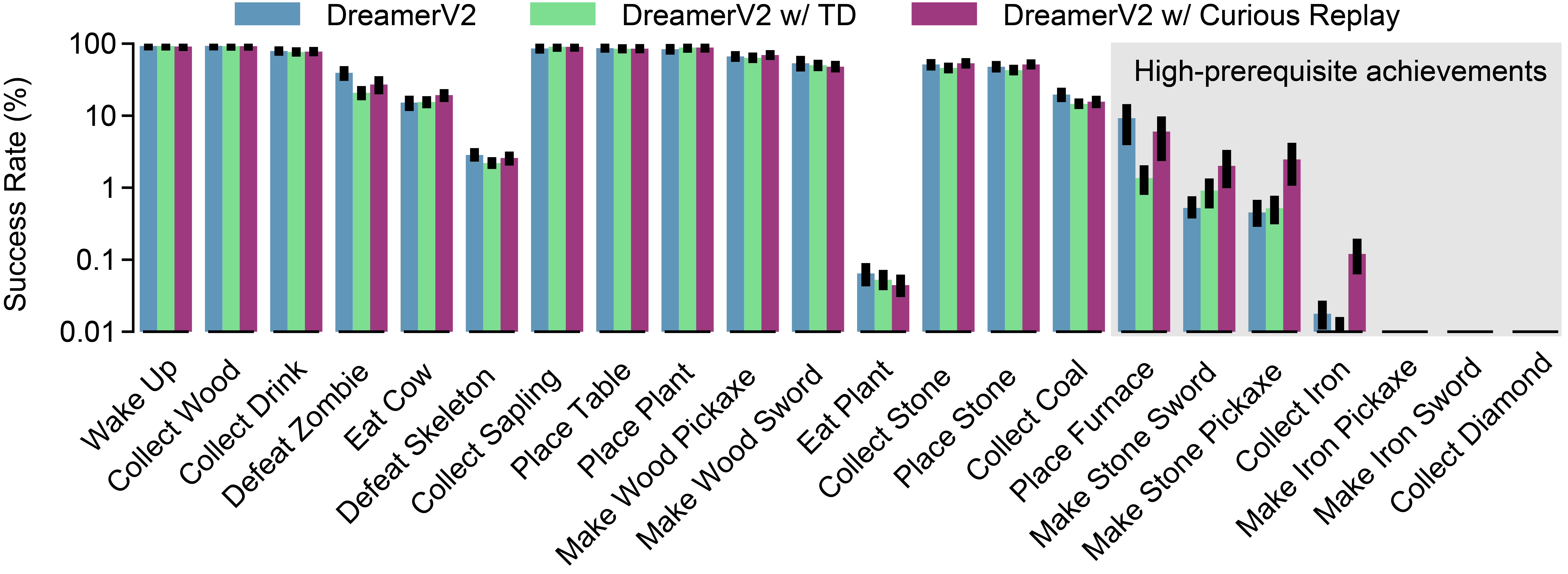}}
\vspace{-10pt}
\caption{DreamerV2 agent ability spectrum showing the success rates for supervised Crafter, ordered from left to right by number of prerequisites for an achievement (n=8 each condition, mean $\pm$ s.e.m.). Curious Replay succeeds at challenging achievements that are unlocked only by many prerequisite. Highlighted high-prerequisite achievements need at least four preceding achievements.}
\label{crafter_spectrum}
\end{center}
\vskip -0.3in
\end{figure*}

\begin{table}[h]
\caption{In Crafter, Curious Replay increases the sampling probability of the steps in the replay buffer where recently discovered resources are held by the agent. This is measured by calculating the median sampling probability for every step where a resource is held relative to the sampling probability in a uniform distribution. The probabilities used are from a saved checkpoint, chosen to be the next multiple of 100k steps after the 1\% achievement threshold is reached (see Table \ref{achievement_progression}). This threshold is intended to represent a small amount of experience with the resource but before the agent may be fully proficient in its use. (n=10 seeds for DreamerV2 w/ Curious Replay, n=5 seeds for DreamerV3 w/ Curious Replay). }
\begin{center}

\begin{NiceTabular}{| l | c  c| c c |}
\toprule
\textbf{Resource} & \textbf{Probability: DreamerV2 w/ CR} & \textbf{Checkpoint} & \textbf{Probability: DreamerV3 w/ CR} & \textbf{Checkpoint} \\
\noalign{\smallskip}\hline\noalign{\smallskip}
Wood & 1.2x & 100k & 1.0x & 100k \\
Wood Pickaxe & 1.7x & 100k & 1.1x & 100k \\
Stone & 1.9x & 100k & 1.2x & 100k\\
Stone Pickaxe & 2.6x & 300k & 1.1x & 300k\\
Iron & 3.1x & 900k & 1.2x & 800k \\
\bottomrule
\end{NiceTabular}
\end{center}
\label{crafter_sampling_prob}
\end{table}

\begin{table}[h]
\caption{Curious Replay increases the number of times steps are trained on when the agent is holding a recently discovered resource. We assessed this for DreamerV3 w/ Curious Replay. This is measured by dividing 1) the number of times each data point with a given resource is trained on by 2) the estimated number of times that each data point's position in the replay buffer would be trained on with a uniform distribution. The checkpoint chosen for each resource is selected to be the nearest multiple of 10k environment steps after the first encounter with the resource. This checkpoint is chosen to give the most granular view of the impact of Curious Replay on new experiences. (n=5 seeds for DreamerV3 w/ Curious Replay). }
\begin{center}

\begin{NiceTabular}{| l | c |}
\toprule
\textbf{Resource} & \textbf{Relative training count} \\
\noalign{\smallskip}\hline\noalign{\smallskip}
Wood & 1.1x \\
Wood Pickaxe & 1.9x \\
Stone & 5.5x \\
Stone Pickaxe & 4.6x\\
Iron & 6.2x \\

\bottomrule
\end{NiceTabular}
\end{center}
\label{crafter_training_count}
\end{table}

\begin{table}[h]
\caption{Steps to 1\% achievement threshold in Crafter (20K step moving average) (n=10 seeds each, 4 for DreamerV3)}
\begin{center}

\begin{NiceTabular}{| l | c  c c c |}
\toprule

\textbf{Achievement} & DreamerV2 & DreamerV2 w/ CR & DreamerV3 & DreamerV3 w/ CR\\
\noalign{\smallskip}\hline\noalign{\smallskip}
Collect Wood & 1K & 1.5K & 0.5K & 0.5K \\
Place Table & 7K & 8K & 4K & \textbf{3K} \\
Make Wood Pickaxe & 46K & 47K & 42K & \textbf{26K} \\
Collect Stone & 131K & 97K & 110K & \textbf{43K} \\
Make Stone Pickaxe & 533K & 236K & 745K & 244K \\
Collect Iron & never & 869K & 973K & \textbf{710K} \\

\bottomrule
\end{NiceTabular}
\end{center}
\label{achievement_progression}
\end{table}

\begin{figure}[h]
\vspace{-15pt}
\vskip 0.2in
\begin{center}
\centerline{\includegraphics[width=\textwidth]{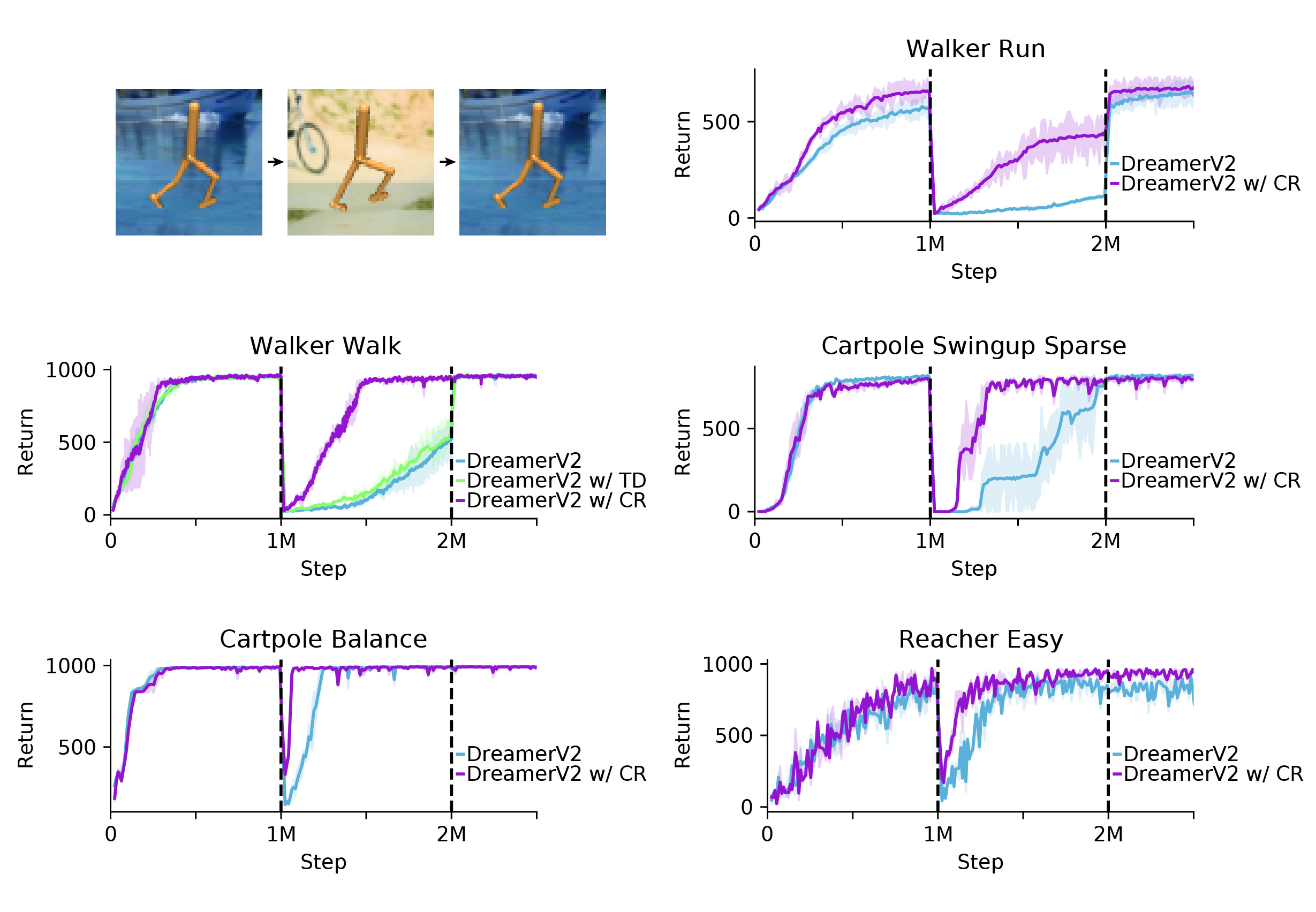}}
\vspace{-15pt}
\caption{Background-Swap Control Suite, comparing DreamerV2 and DreamerV2 with Curious Replay. As schematized in the top left, the background image changes at 1M steps, and reverts at 2M steps. This tests adaptation at step 1M, and also assesses the presence (or, as we observe, absence) of catastrophic forgetting at step 2M (n=3 seeds, mean 
 s.e.m.).}
\label{ddmc_supp}
\end{center}
\vskip -0.4in
\end{figure}

\begin{table}
\caption{Background-Swap Control Suite shows no evidence of catastrophic forgetting, as indicated by the similarly high score between DreamerV2 and DreamerV2 w/ CR at 2.05M steps, right after the background image has reverted to the original image. (CR = Curious Replay, $n$=3 seeds per method, mean $\pm$ s.e.m.) }
\begin{center}

\begin{NiceTabular}{| l | c  c  c  c  c |}

\toprule

\textbf{Method} & \multicolumn{1}{c}{\textbf{Walker Walk}} & \multicolumn{1}{c}{\textbf{Cartpole Swingup Sparse}} & \multicolumn{1}{c}{\textbf{Cartpole Balance}} & \multicolumn{1}{c}{\textbf{Walker Run}} & \multicolumn{1}{c}{\textbf{Reacher Easy}}    \\
\noalign{\smallskip}\hline\noalign{\smallskip}

t=50K steps\\
DreamerV2 & 164.2 $\pm$ 25.4	&2.9 $\pm$ 1.0	&313.1 $\pm$ 45.7	&62.5 $\pm$ 3.8	&68.8 $\pm$ 29.1 \\
DreamerV2 w/ CR & 194.6 $\pm$  32.2	&1.1 $\pm$  1.1	&327.6 $\pm$  23.6	&71.8 $\pm$  17.8	&23.5 $\pm$  19.3 \\
\noalign{\smallskip}\hline\noalign{\smallskip}

t=990K steps\\
DreamerV2 & 954.6 $\pm$ 4.0	&812.3 $\pm$ 2.5	&989.9 $\pm$ 1.1	&578.3 $\pm$ 55.5	&849.9 $\pm$ 92.0 \\
DreamerV2 w/ CR & 956.3 $\pm$ 6.5	&787.1 $\pm$ 13.6	&991.8 $\pm$ 0.4	&655.9 $\pm$ 56.7	&881.2 $\pm$ 59.4 \\
\noalign{\smallskip}\hline\noalign{\smallskip}

t=2.05M steps\\
DreamerV2 & 948.0 $\pm$ 3.9	&817.6 $\pm$ 0.8	&988.8 $\pm$ 4.1	&588.8 $\pm$ 52.2	&808.0 $\pm$ 33.9 \\
DreamerV2 w/ CR & 953.8 $\pm$ 7.8	&804.3 $\pm$ 16.9	&990.3 $\pm$ 1.6	&657.4 $\pm$ 59.7	&943.1 $\pm$ 12.9 \\

\bottomrule
\end{NiceTabular}
\end{center}
\label{ddmc_table}
\end{table}

\begin{figure}[h]
\vspace{-15pt}
\vskip 0.2in
\begin{center}
\centerline{\includegraphics[width=0.3\textwidth]{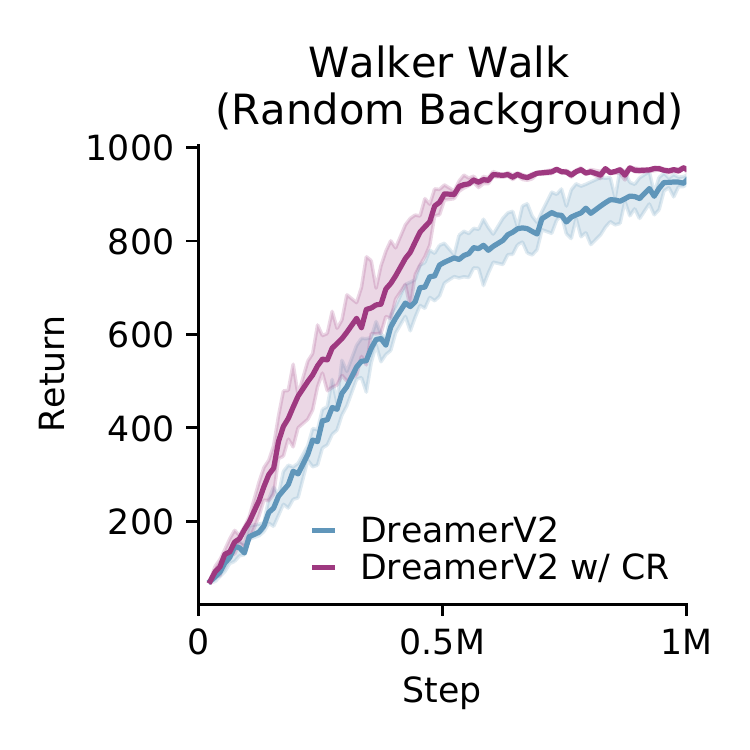}}
\vspace{-15pt}
\caption{Walker Walk task from Distracting Deepmind Control Suite, with a background that changes to a different randomly selected image at each timestep, comparing DreamerV2 and DreamerV2 with Curious Replay (n=5 per task, mean $\pm$ s.e.m.).}
\label{walker_random}
\end{center}
\vskip -0.4in
\end{figure}

\begin{table}[h]
\caption{Deepmind Control Suite, (mean across n=3 seeds per task for DreamerV3 CR, and n=2 seeds each for DreamerV2 CR and DreamerV2). Scores $>50$ more than their counterpart are bolded. $^\dagger$=\mbox{Published  results, see \cite{hafner2023mastering}.}}
\vskip -1in

\begin{center}
\begin{small}
\begin{NiceTabular}{|l|c c| c c |}
\toprule
\textbf{Task} & \textbf{DreamerV3 w/ CR} & \textbf{DreamerV3$^\dagger$} & \textbf{DreamerV2 w/ CR} & \textbf{DreamerV2}\\ 
\midrule
Acrobot swingup & 172.9 & 210.0 & 36.3 & \textbf{321.5} \\
Cartpole balance & 996.4 & 996.4 & 990.9 & 994.3\\
Cartpole balance sparse & 1000.0 & 1000.0 & 985.6 & 991.5 \\
Cartpole swingup & 858.5 & 819.1 & 801.6 & \textbf{871.0} \\
Cartpole swingup sparse & 346.3 & \textbf{792.9} & 328.7 & \textbf{723.2}\\
Cheetah run & \textbf{852.7} & 728.7 & 578.3 & \textbf{786.4} \\
Cup catch & 969.5 & 957.1 & 958.8 & 960.7\\
Finger spin & 557.9 & \textbf{818.5} & \textbf{552.2} & 357.0 \\
Finger turn easy & \textbf{906.3} & 787.7 & 483.8 & \textbf{752.8} \\
Finger turn hard & 777.6 & 810.8 & 404.5 & \textbf{845.3}\\
Hopper hop & 385.1 & 369.6 & 161.8 & \textbf{253.5}\\
Hopper stand & 936.7 & 900.6 & 543.4 & \textbf{756.7}\\
Pendulum swingup & 566.4 & \textbf{806.3} & 743.5 & 784.8 \\
Quadruped run & \textbf{431.1} & 352.3 & 495.0 & \textbf{549.7}\\
Quadruped walk & \textbf{750.8} & 352.6 & \textbf{880.5} & 549.4 \\
Reacher easy & \textbf{956.5} & 898.9 &  \textbf{977.1} & 915.9\\
Reacher hard & 518.8 & 499.2 & 345.7 & \textbf{498.8}\\
Walker run & 772.8 & 757.8 & \textbf{693.3} & 526.7 \\
Walker stand & 978.5 & 976.7 & 977.0 & 971.4 \\
Walker walk & 961.7 & 955.8 & 933.9 & 904.1 \\

\bottomrule
\end{NiceTabular}
\end{small}
\end{center}
\label{dreamerv3_dmc_tasks}
\vskip -0.1in
\end{table}

\begin{figure}[h]
\vspace{-15pt}
\vskip 0.2in
\begin{center}
\centerline{\includegraphics[width=\textwidth]{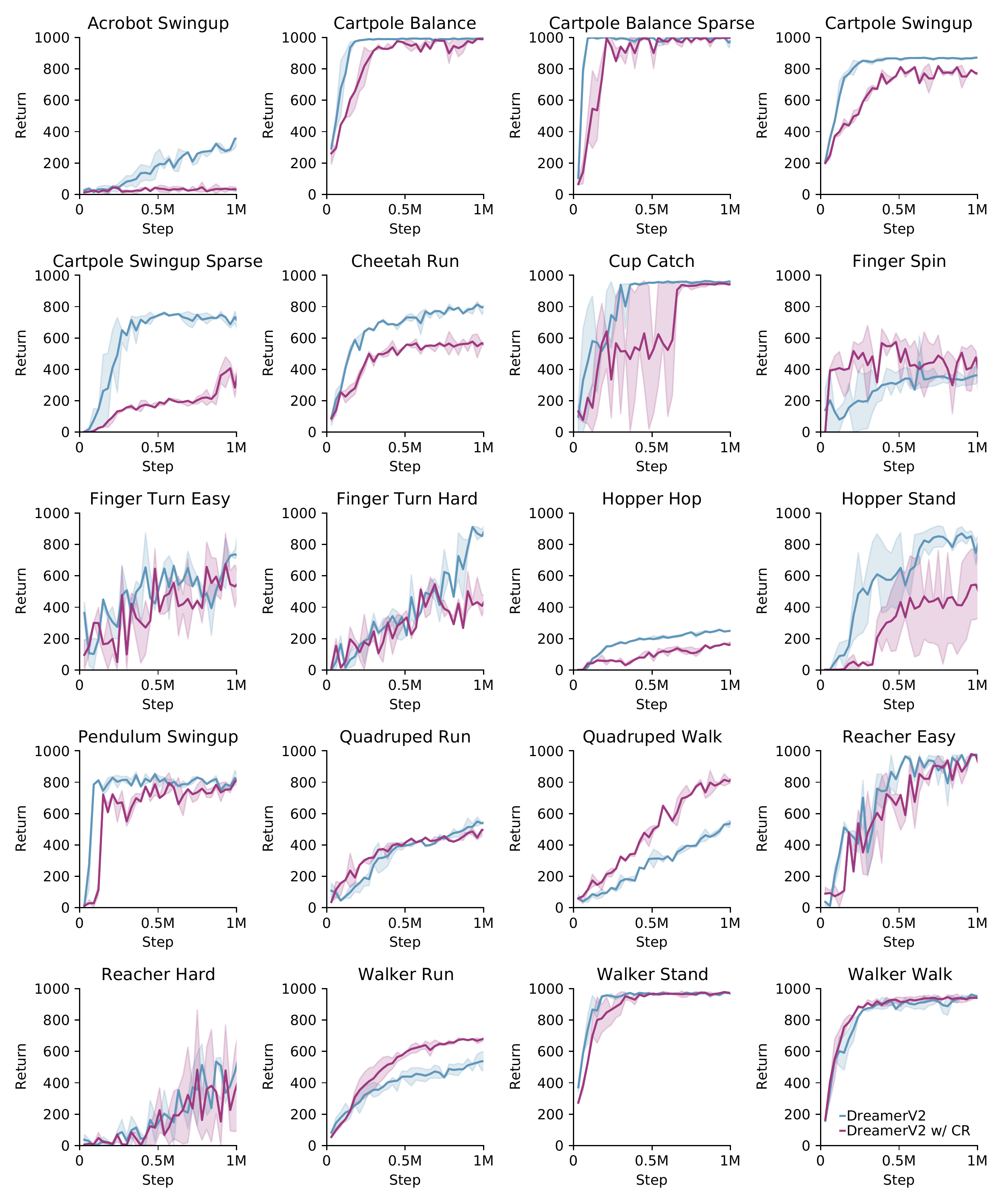}}
\vspace{-15pt}
\caption{Deepmind Control Suite, comparing DreamerV2 and DreamerV2 with Curious Replay. Curious Replay improves performance on tasks such as Quadruped Walk and Walker Run, while decreasing performance on tasks such as Acrobot Swingup and Cartpole Swingup Sparse (n=2 per task, mean $\pm$ s.e.m.).}
\label{dmc_dv2_supp}
\end{center}
\vskip -0.4in
\end{figure}

\begin{figure}[h]
\vspace{-15pt}
\vskip 0.2in
\begin{center}
\centerline{\includegraphics[width=\textwidth]{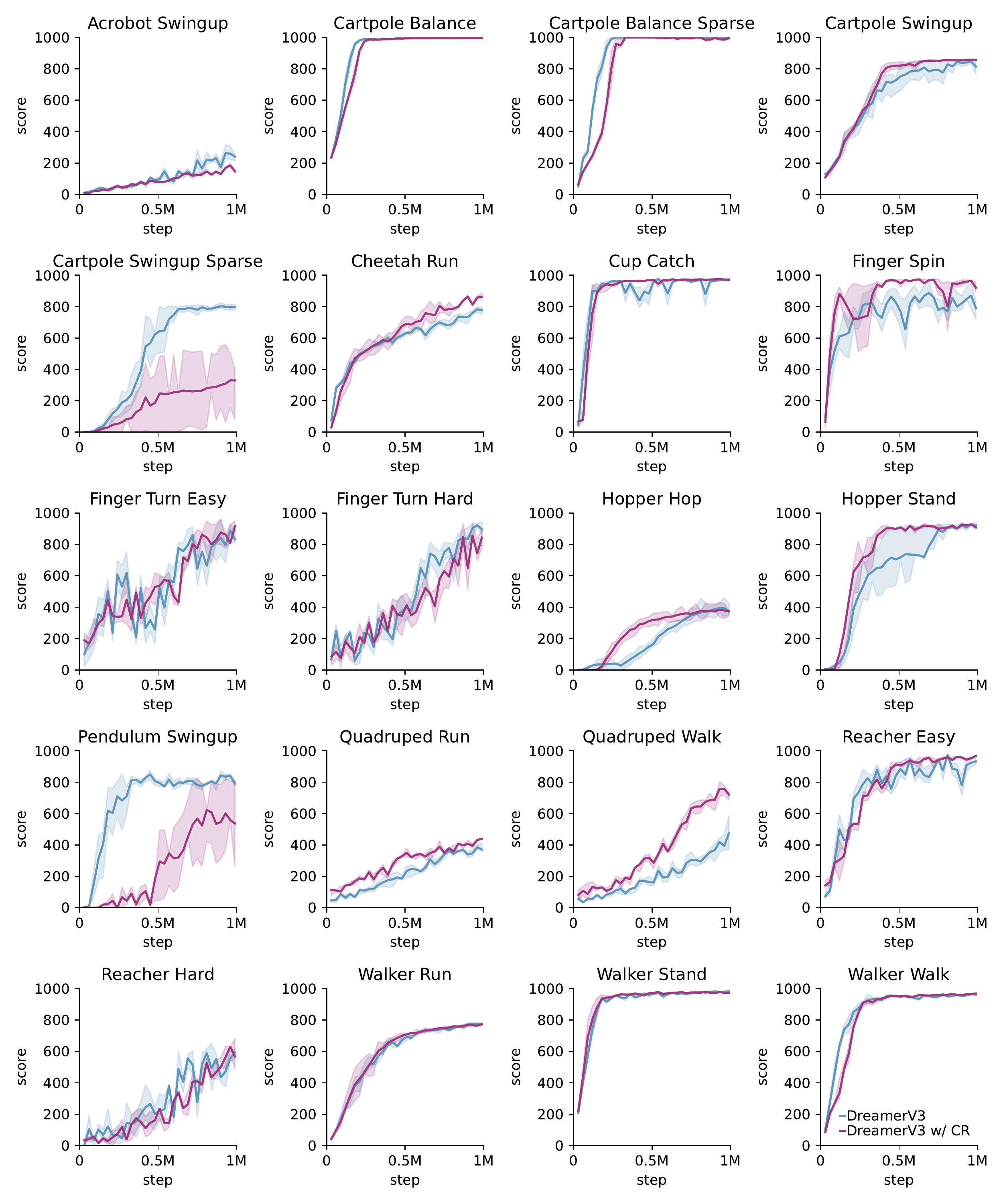}}
\vspace{-15pt}
\caption{Deepmind Control Suite, comparing DreamerV3 and DreamerV3 with Curious Replay. Curious Replay improves performance on tasks such as Quadruped Walk, while decreasing performance on tasks such as Cartpole Swingup Sparse (n=3 per task, mean $\pm$ s.e.m.).}
\label{dmc_dv3_supp}
\end{center}
\vskip -0.4in
\end{figure}

%%%%%%%%%%%%%%%%%%%%%%%%%%%%%%%%%%%%%%%%%%%%%%%%%%%%%%%%%%%%%%%%%%%%%%%%%%%%%%%
%%%%%%%%%%%%%%%%%%%%%%%%%%%%%%%%%%%%%%%%%%%%%%%%%%%%%%%%%%%%%%%%%%%%%%%%%%%%%%%

\end{document}